%% file: main.tex
\documentclass{article}
\usepackage{iclr2027_conference_arxiv,times}
\input{math_commands.tex}

\usepackage{hyperref}
\hypersetup{hidelinks}
\usepackage{url}
\usepackage{graphicx}
\usepackage{booktabs}
\usepackage{tabularx}
\usepackage{colortbl}

\definecolor{tablegroup}{gray}{0.91}
\definecolor{tablehead}{gray}{0.96}
\usepackage{array}
\usepackage{amsmath,amssymb}
\usepackage{xcolor}
\usepackage{microtype}
\usepackage{enumitem}
\usepackage{multirow}
\usepackage{makecell}
\usepackage{pifont}
\usepackage{xspace}
\usepackage{subcaption}
\usepackage{float}

\graphicspath{{figures/}}
\input{generated/numbers.tex}




\input{generated/c-section-numbers.tex}

\definecolor{GeneralBand}{HTML}{E8EEF9}
\definecolor{BaselineBand}{HTML}{F3F4F6}
\definecolor{MechanismBand}{HTML}{EAF4FA}
\definecolor{InterventionBand}{HTML}{E8F6EE}
\definecolor{InfoBand}{HTML}{F1ECF8}
\definecolor{PendingGray}{HTML}{7B8188}

\title{From Hindsight to Foresight: Dissecting Clinical Evidence Acquisition in Large Language Models}
\author{
Ziyang Ling\\
\normalfont School of Biomedical Engineering,\\
\normalfont Division of Life Sciences and Medicine,\\
\normalfont University of Science and Technology of China\\
\normalfont Hefei, Anhui, China\\
\normalfont \texttt{zaneling@mail.ustc.edu.cn}
}

\begin{document}
\maketitle

\begin{abstract}
Large language models (LLMs) often diagnose full sight, static clinical cases well yet struggle in foresight, dynamic diagnosis.
Prior work has improved this limitation by quantifying the information models acquire and by developing stronger follow-up strategies to improve diagnostic accuracy.
In contrast, we introduce a distinctive hindsight to systematically dissect why LLMs fail to acquire diagnostically critical evidence, tracing the failure from evidence identification, utilization and downstream diagnostic updating through not only diagnostic accuracy, but also evidence coverage, trajectory dynamics, and elicited probability and information entropy.
Starting experiments from the breakdown at the level of capabilities, to the diagnosis of the pre-event state before the evidence arrives, and to the post-event state transition after the evidence arrives, our research has revealed the profound limitations and internal causes existing in the dynamic diagnostic process.
At the same time, we have also conducted sufficient verification and generalization experiments to prove that these limitations are widespread. Together, these findings shift the central challenge from whether LLMs possess relevant clinical knowledge to whether they can identify, acquire, and effectively use the right evidence under clinical diagnosis and conduct a more detailed mechanism analysis to identify the exact location of the capability gap.
\end{abstract}

\section{Introduction}

\begin{figure}[t]
    \centering
    \includegraphics[width=\linewidth]{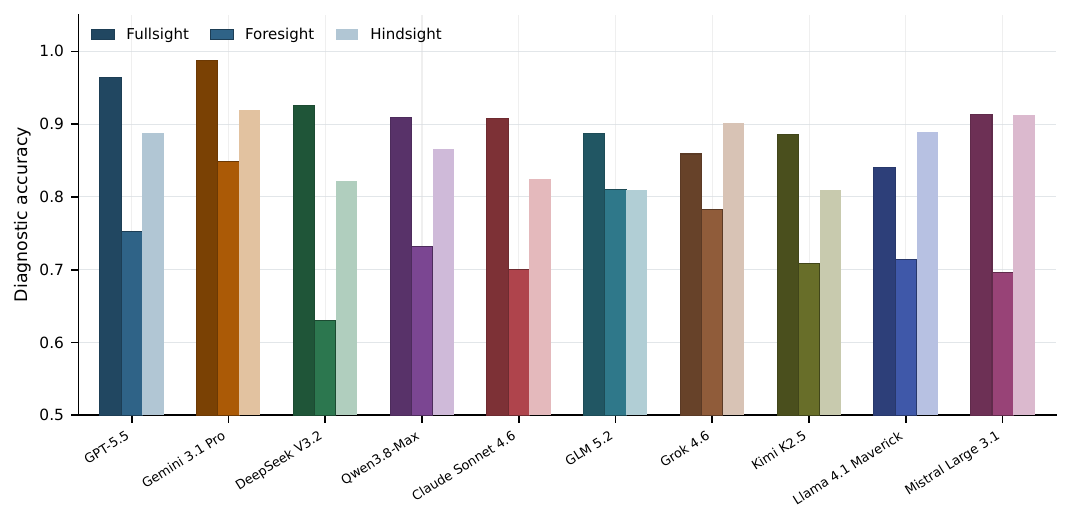}
    \caption{
Diagnostic accuracy of ten general-purpose LLMs under three evidence-access settings
(798 model--case pairs).
Fullsight receives the complete case upfront;
Foresight prospectively acquires evidence over at least six turns;
and Hindsight retrospectively selects and orders diagnostically critical evidence
from the completed case without reference diagnosis, before re-diagnosis.
}
    \label{fig:tenmodel}
\end{figure}

Large language models (LLMs) can answer many medical questions, make clinical judgments, and achieve very high diagnostic accuracy when complete clinical evidence is presented upfront \citep{singhal2023clinical,singhal2025expert,arora2025healthbench}. However, such strong performance on static, information-complete tasks does not reliably translate to realistic clinical capabilities, where LLMs still exhibit substantial performance degradation when they must interact with patients, actively acquire missing evidence, and reason through sequential diagnostic processes \citep{bean2026reliability,chen2026grounding,li2024mediq,johri2025craftmd,tu2025amie}. This gap has motivated the development of dynamic and interactive diagnostic benchmarks, as conventional static question-answering benchmarks cannot adequately capture the iterative evidence acquisition and sequential decision-making required in real-world clinical diagnosis \citep{schmidgall2026agentclinic,chen2026grounding}.
Besides, EviMed and related evidence-centered work measure whether relevant or sufficient information is covered; expected-information and evidence-gap methods propose better targets or policies \citep{long2026strong,huang2026mtinfoseek,wong2026medclarify,sanghvi2026medxagent,dai2026cdeg}.

To bridge this capability gap, several recent studies have explored methods to improve follow-up questioning and progressive evidence acquisition. \textsc{FollowupQ} uses a multi-agent framework to generate personalized follow-up questions by jointly leveraging patient messages and longitudinal EHR context \citep{gatto2025followup}. \textsc{KG-Followup} incorporates structured medical knowledge graphs to guide the identification of missing clinical information and generate more comprehensive questions \citep{sun2026linking}. \textsc{DDO} formulates symptom inquiry and diagnosis as separate but collaborative decision processes, explicitly optimizing sequential information acquisition \citep{jia2025ddo}. \textsc{Doctor-R1} uses agentic reinforcement learning to learn high-yield inquiry strategies over multi-turn clinical interactions \citep{lai2026doctorr1}.

Benchmarks and studies show that prospective acquisition is difficult and that stronger evidence collection can help, but they do not jointly trace the behavioral mechanism from an attempted action to the evidence actually obtained and then to diagnosis. Rather than treating this gap as a surface performance deficit, we seek to dissect the underlying behavioral process and identify where and why prospective evidence acquisition breaks down. Our key question is therefore narrower and deeper: \emph{what specifically limits LLMs' prospective evidence acquisition, which factors most strongly shape these limitations, and what underlying mechanisms give rise to them?}

Most existing static evaluations contrast complete-information disclosure once, what we call \textbf{Fullsight}, with prospective interactive diagnosis what we call \textbf{Foresight} that models perform well when all evidence is available upfront. MedAction analyzes multi-turn diagnosis and identifies ungrounded test ordering, unreliable diagnostic updating, and degraded conversational coherence as recurring failures \citep{hsu2026medaction}. CDEG compares successful and failed diagnostic trajectories to identify decision-critical evidence, then tests its effect through controlled interventions \citep{dai2026cdeg}.

However, this contrast alone cannot show whether models can recognize useful evidence once a case is complete, or where that evidence is lost during a step-by-step consultation. We introduce \textbf{Hindsight} as a retrospective perspective: after seeing the fully realized case, the model identifies the facts most useful for diagnosis. This reveals something Fullsight accuracy cannot show on its own that a model may recognize useful evidence after seeing its result but fail to seek it actively and priority while the result is unknown. We first observe this pattern across ten completed models shown in Figure~\ref{fig:tenmodel}. We then compare Hindsight and Foresight at each disclosure depth, tracing what the model asks, what evidence it obtains, and how its diagnosis changes.
Our focus of analysis can be presented as
\begin{equation}
\underbrace{A_{\mathrm{Full}}-A_{\mathrm{Fore}}}_{\substack{16.8\,\mathrm{pp}\ \text{total gap}\\\textbf{Phenomenon}}}
=
\underbrace{(A_{\mathrm{Full}}-A_{\mathrm{Hind}})}_{\substack{4.5\,\mathrm{pp}\\\text{limited loss}}}
+
\underbrace{(A_{\mathrm{Hind}}-A_{\mathrm{Fore}})}_{\substack{12.3\,\mathrm{pp}\\\textbf{focus of our analysis}}}.
\label{eq:hindsight-decomposition}
\end{equation}
The small first difference shows that retrospectively selected evidence retains much of the diagnostic value of the complete case. The second difference motivates our deeper analysis of clinical evidence acquisition ability gap in interactive diagnosis.

Our contributions are: (1) We introduce hindsight as a new perspective to examine evidence recognition in interactive diagnosis that LLMs often fail to find those facts. Fullsight demonstrates LLMs do not lack medical knowledge and hindsight proves evidence recognition ability in LLMs and provides a reference for the cognitive ability limit which reveals a gap between recognizing useful evidence afterward and acquiring it while the outcome is still unknown; (2) We examine this gap throughout the foresight trajectory. At each step, we track what the model asks, what new evidence it actually obtains, and how its diagnosis changes. Comparing these trajectories with the useful facts identified in hindsight lets us distinguish problems that arise before evidence arrives from problems that persist after it arrives. Controlled evidence additions then test whether a missed fact can change the diagnosis; (3) We test the findings from multiple facets. Comparisons of question formats, fact counts, evidence order, and diagnostic history help rule out simple explanations for the gap. Further tests across models and datasets examine how broadly the pattern holds and prove the prevalence of the conclusions drawn from our research.

\section{Related Work}
\paragraph{Interactive diagnosis benchmarks.}
Recent benchmarks have moved beyond static medical QA toward interactive and sequential diagnosis. MediQ, CRAFT-MD, and AMIE evaluate adaptive questioning and simulated diagnostic conversations \citep{li2024mediq,johri2025craftmd,tu2025amie}; AgentClinic, AI Hospital, and MedAgentBench provide simulated clinical or EHR environments \citep{schmidgall2026agentclinic,fan2025aihospital,jiang2025medagentbench}. MedR-Bench, Q4Dx, MINT, DiagnosisArena, RareArena, and CPC-Bench further extend evaluation to test selection, sequential revision, and broader diagnostic reasoning \citep{qiu2025medrbench,werthaim2026q4dx,fang2026mint,zhu2026diagnosisarena,chen2026rarearena,buckley2025century}.

\paragraph{Follow-up methods.}
A growing line of work improves how models decide what information to acquire next and how the acquired evidence shapes their judgments. EviMed and MT-InfoSeek explicitly model missing or sufficient evidence \citep{long2026strong,huang2026mtinfoseek}; MedClarify, MeDxAgent, and CDEG prioritize diagnostically informative or decision-critical evidence \citep{wong2026medclarify,sanghvi2026medxagent,dai2026cdeg}. Sequential Diagnosis and Doctor-R1 instead learn cost-aware or adaptive inquiry policies over multi-turn interactions \citep{nori2025sequential,lai2025doctorr1}. Evidence-auditing studies examine output changes under controlled evidence subsets \citep{liao2026auditing}, information-based approaches analyze how particular evidence shifts diagnostic uncertainty, and hindsight studies show that access to outcome or future information can alter subsequent judgments \citep{fischhoff1975hindsight,matsuura2026hindsight}.

\section{Experimental Setup}
\label{sec:design}

\subsection{Experimental Design}
\label{sec:experimental-design}

\begin{figure}[t]
    \centering
    \includegraphics[width=\linewidth]{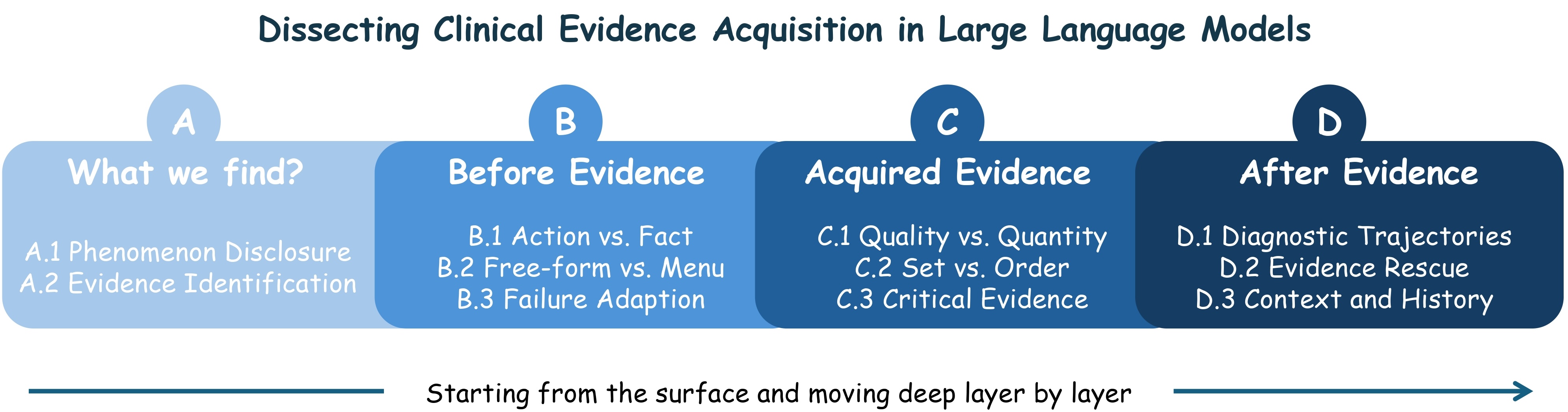}
    \caption{Experiment design overview. Starting from the observed ability gap (A) to three steps of clinical evidence acquisition of LLMs, including (B) before evidence, (C) acquired evidence analysis and (D) after evidence examination.}
    \label{fig:exdegisn}
\end{figure}

Figure~\ref{fig:exdegisn} organizes the study from the visible performance gap to the stages at which it arises. We first establish the contrast among fullsight diagnosis, foresight acquisition, and hindsight retrospective evidence identification (A). Results in Figure~\ref{fig:tenmodel} reveal the phenomenon. Then, we delved deeper into the phenomenon, and thoroughly analyzed where the gap in evidence acquisition lies.
We divide evidence acquisition into three steps.
First, before a result is known (B), we distinguish requests from newly obtained evidence, compare free-form requests with a result-blind  menu, and examine adaptation after no-match and duplication actions. Second, for the evidence actually acquired (C), we match evidence counts case by case, separate evidence-set effects from order effects, and test whether diagnostically critical facts are reached. Third, after evidence arrives (D), we track diagnostic transitions, intervene with a retrospectively identified fact, and test the entropy presentation, diagnostic distribution and how prior context history affect the response. 
The final part E includes validation experiments for recurrence, ablation, and robustness.

\subsection{Datasets and Cohorts.}
The frozen source universe contains \SourceN{} cases: \SourceAgentClinicN{} AgentClinic MedQA cases, \SourceDermN{} dermatology cases, and \SourceDiagnosisArenaN{} DiagnosisArena cases. Thus the primary estimand is conditional on complete-case competence, not population-level clinical accuracy. Prespecified derived cohorts are the result-blind menu cohort ($N=\MenuN$), evidence-set/order cohort ($N=\OrderN$), selector-robustness cohort ($N=\SelectorN$), four-model shared cohort ($N=\XModelN$), shared menu cohort ($N=\SharedMenuN$), and balanced EviMed bridge ($N=\EviN$). The generated cohort-flow file records every case ID.

\subsection{Models and Configuration.}
Initial experiment completed \BroadCompleted{} general LLMs, yielding \BroadPairs{} eligible model--case pairs. The strict study uses GPT-5.5 and separates four roles: a selector proposes evidence targets, the environment matcher realizes at most one canonical fact, a fresh diagnostic call reads each evidence prefix, and a diagnosis-equivalence judge scores the output. The shared replications additionally use DeepSeek V3.2, Gemini 3.1 Pro Preview, and GLM 5.2. We use OpenRouter as API service, temperature 0, provider-default top-$p$, maximum outputs of 3,500 tokens by stage, and at most three transport retries.

\subsection{Metrics}
\label{sec:metrics}

Common metrics include Diagnostic Accuracy, Evidence Yield, Critical-Evidence Reach, Stable Sufficiency, Wrong-State Persistence, and Information Coverage Rate. Individual experiments use additional task-specific metrics. Owing to space constraints, all metric definitions, formulas, and denominators are provided in Appendix.

\section{Results and Analysis}
\label{sec:experiments}

\input{sections/results/A-ability-gap.tex}

\input{sections/results/B-before-evidence.tex}
\input{sections/results/C-acquired-evidence.tex}
\input{sections/results/D-after-evidence.tex}

\input{sections/results/E-robustness-transfer.tex}

\section{Conclusion}
\label{sec:conclusion}

Our work reframes the gap between static and interactive diagnosis as a
failure to translate recognized evidence value into prospective action.
Fullsight establishes that the model can solve the completed case, while
hindsight shows that, once the case is realized, the same model can identify a compact and diagnostically useful trajectory without access to the reference diagnosis. The persistent separation from foresight therefore cannot be reduced to missing disease knowledge. Our stage-wise analysis localizes the loss: (1) before evidence arrives, requests do not reliably obtain novel facts, and it is hard to recover from previous failure; (2) among the facts obtained, identity and order matter beyond quantity, while distributional change or increased confidence
does not guarantee diagnostic truth-directed progress; (3) after evidence arrives, high-value facts can produce earlier updates and directly rescue incorrect diagnoses. Recurrence across model backbones and data sources supports a multi-stage evidence acquisition bottleneck.

Hindsight is neither a deployable acquisition policy nor evidence that the
entire retrospective gap is avoidable. It is an empirical reference that
reveals what the model's existing knowledge can support once uncertainty
about evidence realization is removed, and our conclusions remain
conditional on complete-case competence and controlled evidence
environments. The next challenge is to convert this retrospective recognition
into prospective control under the information available at decision time.
Future clinical follow-up should jointly optimize target realizability, novelty, diagnostic value, sequencing, and failure-specific recovery, while being evaluated across the full chain from request to realized evidence to
diagnostic transition. Closing the gap from hindsight to foresight therefore
means turning available medical knowledge into timely, reliable
evidence-seeking behavior.

\subsection*{AI use statement}

During the preparation of this manuscript, the authors used generative AI tools (large language models) only for limited and well-defined tasks: improving the clarity and grammar of the prose, formatting tables and citations, and writing routine code boilerplate such as data-processing and plotting scripts. These tools were not used to generate research ideas, hypotheses, scientific claims, experimental results, metrics, or analyses, nor to survey or summarize the literature without independent verification by the authors. 

\bibliography{references}
\bibliographystyle{iclr2027_conference}

\clearpage

\appendix
\input{sections/appendix.tex}

\end{document}

%% file: math_commands.tex
\usepackage{amsmath,amsfonts,bm}

\def\eqref#1{equation~\ref{#1}}
\def\1{\bm{1}}

\DeclareMathAlphabet{\mathsfit}{\encodingdefault}{\sfdefault}{m}{sl}
\SetMathAlphabet{\mathsfit}{bold}{\encodingdefault}{\sfdefault}{bx}{n}

%% file: generated/numbers.tex
\newcommand{\SourceN}{320}

\newcommand{\SourceAgentClinicN}{140}
\newcommand{\SourceDermN}{80}
\newcommand{\SourceDiagnosisArenaN}{100}

\newcommand{\BroadCompleted}{10}

\newcommand{\BroadPairs}{798}

\newcommand{\FillRate}{58.7\%}
\newcommand{\NoMatchRate}{16.0\%}
\newcommand{\DuplicateRate}{25.3\%}

\newcommand{\OrderN}{100}

\newcommand{\XModelN}{59}
\newcommand{\SharedMenuN}{50}
\newcommand{\SelectorN}{100}
\newcommand{\MenuN}{120}

\newcommand{\MenuFreeFactsThree}{2.09}

\newcommand{\EviN}{200}

%% file: generated/c-section-numbers.tex
\providecommand{\OrderN}{100}

%% file: sections/results/A-ability-gap.tex
\subsection{A. Ability Gap}
\label{sec:results-capability}

\subsubsection{A.1 Phenomenon disclosure}
\label{sec:exp-a1}
Experiments across 10 general LLMs shown in figure~\ref{fig:tenmodel} reveal a V-shaped diagnostic accuracy curve in fullsight--foresight--hindsight trajectory. It is clear that LLMs do not lack knowledge, and they have sufficient ability to review and identify relevant evidence, and can select and prioritize the most diagnostically valuable evidence from the full evidence set which is an upper limit that a model can potentially reach under comprehensive consideration. Obviously, there are still limitations in the ability to acquire evidence in foresight, which have not yet reached the upper limit.

\subsubsection{A.2 Critical Evidence}
Foresight’s poor performance cannot simply be attributed to an inability to diagnose the disease. We examine this gap in the primary cohort (\(N=181\)) and a separate 12-step continuation cohort (\(N=140\); Table~\ref{tab:primary}). At \(k=3\), Foresight reached 33.1\% accuracy, whereas hindsight reached 73.5\%, yielding a paired gap of 40.3 percentage points (95\% CI 32.6--48.6). 
To assess the direction and consistency of the paired difference, we report the matched-pair odds ratio and the two-sided exact McNemar \(p\)-value, both calculated from cases in which only one condition produced the correct diagnosis. At \(k=3\), discordant outcomes strongly favored Hindsight over Foresight (OR \(=15.60\), \(p=6.40\times10^{-18}\)). Extending the original policy to twelve requests improved Foresight to 52.5\%, but Hindsight remained higher at 77.3\%, leaving a 24.9-point gap (95\% CI 17.1--32.0; OR \(=7.43\), \(p=1.36\times10^{-9}\)). The continuation cohort showed the same pattern, demonstrating that the recognition advantage remained directionally consistent even under a larger acquisition budget. Thus, hindsight identifies realized evidence that supports correct diagnosis substantially more often than the evidence acquired prospectively.

We therefore use hindsight as a plausible empirical upper reference for the remainder of the analysis. It is not an existing clinical acquisition method, a deployable policy, or a formal global optimum. Rather, it asks what diagnostic trajectory the same LLM can retrospectively construct from its own medical knowledge once the complete realized case is visible, while withholding the reference diagnosis. In this sense, hindsight approximates the upper envelope of evidence recognition available to the model when uncertainty about which observations will be realized has been removed.

\input{tables/results/A/recognition-acquisition.tex}

%% file: tables/results/A/recognition-acquisition.tex
\begin{table*}[t]
\centering
\caption{Fine-grained recognition–acquisition gaps across evidence depth, comparing retrospective Hindsight with prospective Foresight in two separately analyzed cohorts.}
\label{tab:primary}

\begingroup
\scriptsize
\setlength{\tabcolsep}{2.2pt}
\renewcommand{\arraystretch}{0.94}

\resizebox{\textwidth}{!}{%
\begin{tabular}{
cccccc
@{\hspace{7pt}}
ccccc
}
\toprule

\textbf{Round}
& \multicolumn{5}{c}{\textbf{Panel A: Primary cohort} ($N=181$)}
& \multicolumn{5}{c}{\textbf{Panel B: Continuation cohort} ($N=140$)}
\\[-2pt]

\cmidrule(lr){2-6}
\cmidrule(lr){7-11}

\textbf{$k$}
& \textbf{Foresight}
& \textbf{Hindsight}
& \textbf{Gap [95\% CI]}
& \textbf{OR}
& \textbf{Exact $p$}
& \textbf{Foresight}
& \textbf{Hindsight}
& \textbf{Gap [95\% CI]}
& \textbf{OR}
& \textbf{Exact $p$}
\\

\midrule

1
& 18.8 & 55.8 & 37.0 [29.3, 45.3]
& 12.17 & $1.00\times10^{-15}$
& 15.7 & 57.1 & 41.4 [32.9, 50.0]
& 20.33 & $4.74\times10^{-15}$
\\

2
& 29.3 & 66.3 & 37.0 [28.7, 45.3]
& 9.38 & $9.10\times10^{-15}$
& 25.7 & 66.4 & 40.7 [31.4, 50.0]
& 12.40 & $1.42\times10^{-13}$
\\

3
& 33.1 & 73.5 & 40.3 [32.6, 48.6]
& 15.60 & $6.40\times10^{-18}$
& 29.3 & 70.0 & 40.7 [31.4, 50.0]
& 15.25 & $3.92\times10^{-14}$
\\

4
& 35.9 & 75.1 & 39.2 [31.5, 47.0]
& 18.75 & $5.24\times10^{-18}$
& 30.7 & 75.0 & 44.3 [35.7, 52.9]
& 32.00 & $6.00\times10^{-17}$
\\

5
& 34.8 & 74.0 & 39.2 [30.9, 47.5]
& 9.88 & $8.38\times10^{-16}$
& 28.6 & 73.6 & 45.0 [35.7, 54.3]
& 16.75 & $8.74\times10^{-16}$
\\

6
& 42.0 & 75.7 & 33.7 [26.0, 41.4]
& 11.17 & $3.95\times10^{-14}$
& 37.1 & 73.6 & 36.4 [27.1, 45.0]
& 13.75 & $1.70\times10^{-12}$
\\

7
& 46.4 & 79.6 & 33.1 [26.0, 40.9]
& 16.00 & $5.87\times10^{-15}$
& 42.9 & 77.1 & 34.3 [25.7, 42.9]
& 13.00 & $1.10\times10^{-11}$
\\

8
& 47.0 & 74.6 & 27.6 [19.3, 35.9]
& 5.55 & $1.55\times10^{-9}$
& 41.4 & 72.9 & 31.4 [21.4, 41.4]
& 5.89 & $1.05\times10^{-8}$
\\

9
& 49.7 & 77.3 & 27.6 [19.3, 35.9]
& 6.00 & $8.00\times10^{-10}$
& 45.7 & 77.9 & 32.1 [22.9, 41.4]
& 7.43 & $1.36\times10^{-9}$
\\

10
& 49.7 & 76.8 & 27.1 [19.3, 34.8]
& 6.44 & $6.81\times10^{-10}$
& 45.0 & 76.4 & 31.4 [22.1, 40.7]
& 7.29 & $2.40\times10^{-9}$
\\

11
& 53.6 & 77.3 & 23.8 [16.0, 31.5]
& 6.38 & $9.05\times10^{-9}$
& 49.3 & 77.1 & 27.9 [18.6, 37.1]
& 7.50 & $1.83\times10^{-8}$
\\

12
& 52.5 & 77.3 & 24.9 [17.1, 32.0]
& 7.43 & $1.36\times10^{-9}$
& 47.9 & 77.1 & 29.3 [20.0, 37.9]
& 7.83 & $5.81\times10^{-9}$
\\

\bottomrule
\end{tabular}%
}

\endgroup
\end{table*}

%% file: sections/results/B-before-evidence.tex
\subsection{B. Before Evidence: Target Choice and Action Realization}
\label{sec:results-action}

Having established hindsight as a retrospective upper limit reference, we first
localize the gap before evidence reaches the diagnostic step. Foresight evidence acquisition can fail because the model selects an uninformative target or
because the requested information is not successfully realized.

\subsubsection{B.1 Request $\neq$ Evidence}
\label{sec:exp-b1}

An evidence request does not translate one-to-one into new evidence.
Only \FillRate{} of request slots yield a new unique fact in Figure~\ref{fig:realization}.
\DuplicateRate{} repeat an observed fact and \NoMatchRate{} return no
matched evidence. Consequently, three free-form requests produce only
\MenuFreeFactsThree{} unique facts on average. The effective evidence
budget is therefore substantially smaller than the nominal request budget.

\subsubsection{B.2 Menu Vs. Freeform}
\label{sec:exp-b2}

\begin{figure}[t]
    \centering
    \includegraphics[width=\linewidth]{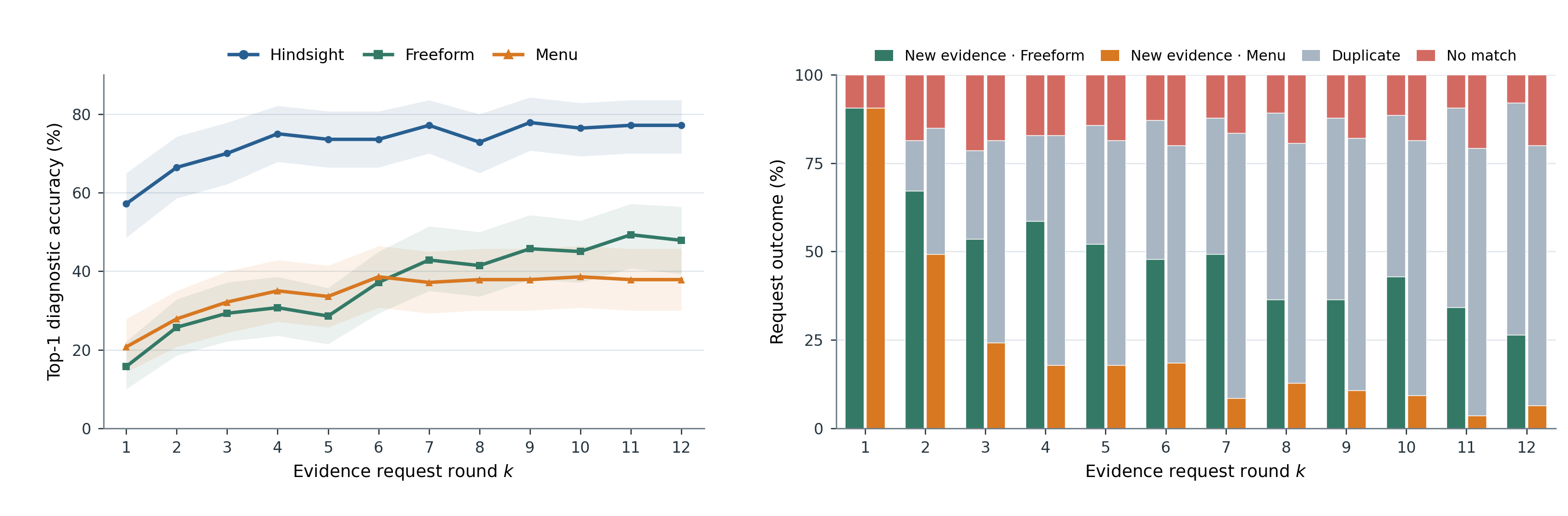}
    \caption{Acquired evidence analysis.
    Left: Diagnostic accuracy across request rounds under
    hindsight, free-form Foresight, and result-blind menu foresight;
    Right: Outcomes of free-form and menu requests, partitioned
    into new evidence, duplicate evidence, and no match.}
    \label{fig:realization}
\end{figure}

Free-form foresight may fail because a natural-language request does not match a canonical fact or targets information that is absent from the realized case. To test whether these interface-level failures explain the acquisition gap, we introduce a result-blind menu that replaces open-ended request generation with a fixed set of candidate targets while withholding their outcomes. Shown in the right part of Figure~\ref{fig:realization}, menu and free-form
Foresight both reach 31.7\% accuracy at \(k=3\), substantially below the
75.0\% achieved by hindsight. Thus, the gap cannot be explained solely by free-text mismatch or requests for unavailable evidence: even with a structured target space, the model still struggles to select and realize diagnostically informative evidence.

\subsubsection{B.3 Adaptation after failure}
\label{sec:exp-b3}

\begin{figure}[t]
    \centering
    \includegraphics[width=\linewidth]{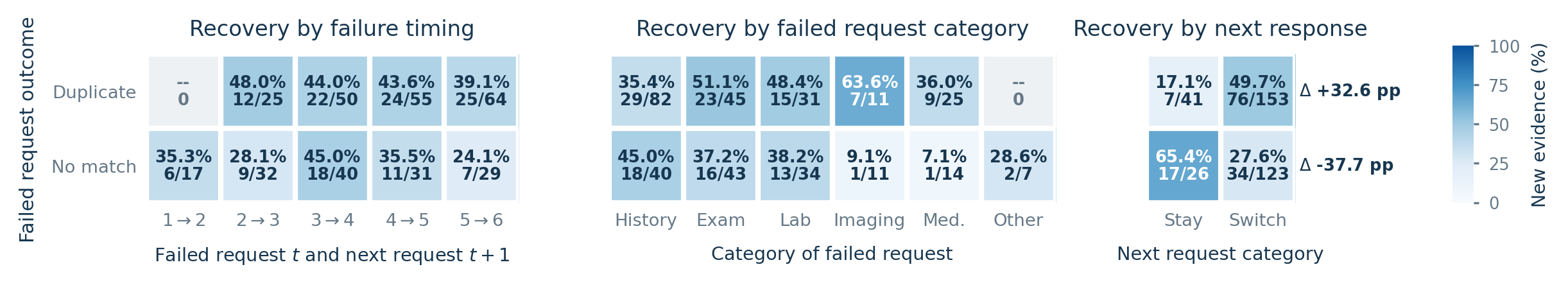}
    \caption{Recovery after unsuccessful evidence requests.
Next-request recovery rates are stratified by failure timing, failed-request category, and whether the next request stays in or switches category.}
    \label{fig:b3-adaptation}
\end{figure}

We then examine whether foresight recovers after a request produces no new
evidence. Among 343 duplicate or no-match events with an eligible subsequent
request, the next request obtains new evidence after 83/194 duplicates
(42.8\%) and 51/149 no-match outcomes (34.2\%). Recovery varies across
failure positions and request categories rather than improving monotonically
with additional turns (Figure~\ref{fig:b3-adaptation}).

The value of changing request category depends on the preceding failure.
After a duplicate, switching category is associated with higher recovery
than staying in the same category (49.7\% versus 17.1\%). After a no-match,
the association reverses: recovery is lower after switching than after
staying (27.6\% versus 65.4\%). The acquisition bottleneck arises not only from target selection and action realization, but also from failure-specific adaptation across successive requests.

%% file: sections/results/C-acquired-evidence.tex
\subsection{C. Acquired Evidence: Composition and Diagnostic Value}
\label{sec:results-evidence}

As we have previously evaluated the relation between request and evidence, now we focus on the evidence that entered the
case state and ask two questions: what determines its diagnostic value, and
whether information change is sufficient to identify the diagnostic value.

\input{tables/results/C/evidence-value-decomposition.tex}
\subsubsection{C.1 Evidence amount, identity, and order}
\label{sec:exp-c1}

As shown in Table~\ref{tab:evidence-value-decomposition}, we first test whether the gap is explained by foresight acquiring fewer evidence quantity. For each case and request budget, we truncate hindsight to the number of unique facts actually obtained by Foresight. At $k=3$, foresight has acquired 2.09 facts on average and reaches 33.1\% accuracy, whereas count-matched hindsight reaches 63.0\%, leaving a paired gap of 29.8 percentage points. The gap remains 28.7 points at $k=6$ (42.0\% versus 70.7\%). In the separate continuation cohort, a 21.4-point gap remains at $k=12$. Thus, acquiring fewer facts explains only part of the foresight deficit.

The remaining difference is also associated with evidence identity. Critical
Evidence Recall is only 16.0\% at $k=3$, and the hindsight Top-1 fact is never reached within six requests in 64.6\% of primary cases. Meanwhile,
hindsight reaches stable diagnostic correctness earlier: at $k=2$, the
stable-correct rates are 50.3\% for hindsight and 23.2\% for foresight. Foresight therefore often accumulates evidence without reaching the facts prioritized by the retrospective reference.

Direct interventions further separate evidence identity from order. In the
fixed $N=100$ panel, reranking the Foresight set raises normalized
six-prefix accuracy from 34.3\% to 43.7\%, a gain of 9.3 points. Replacing that set with the selected set under the
same ranking rule raises accuracy to 70.3\%, a further gain of 26.7 points
. Reversing the Hindsight order reduces accuracy from 70.3\%
to 44.3\%, an order effect of 26.0 points.
Evidence identity and order therefore make distinct contributions:
selection determines which diagnostic value is available, while order
determines how early that value becomes accessible.

\begin{figure*}[t]
\centering
\includegraphics[width=\textwidth]{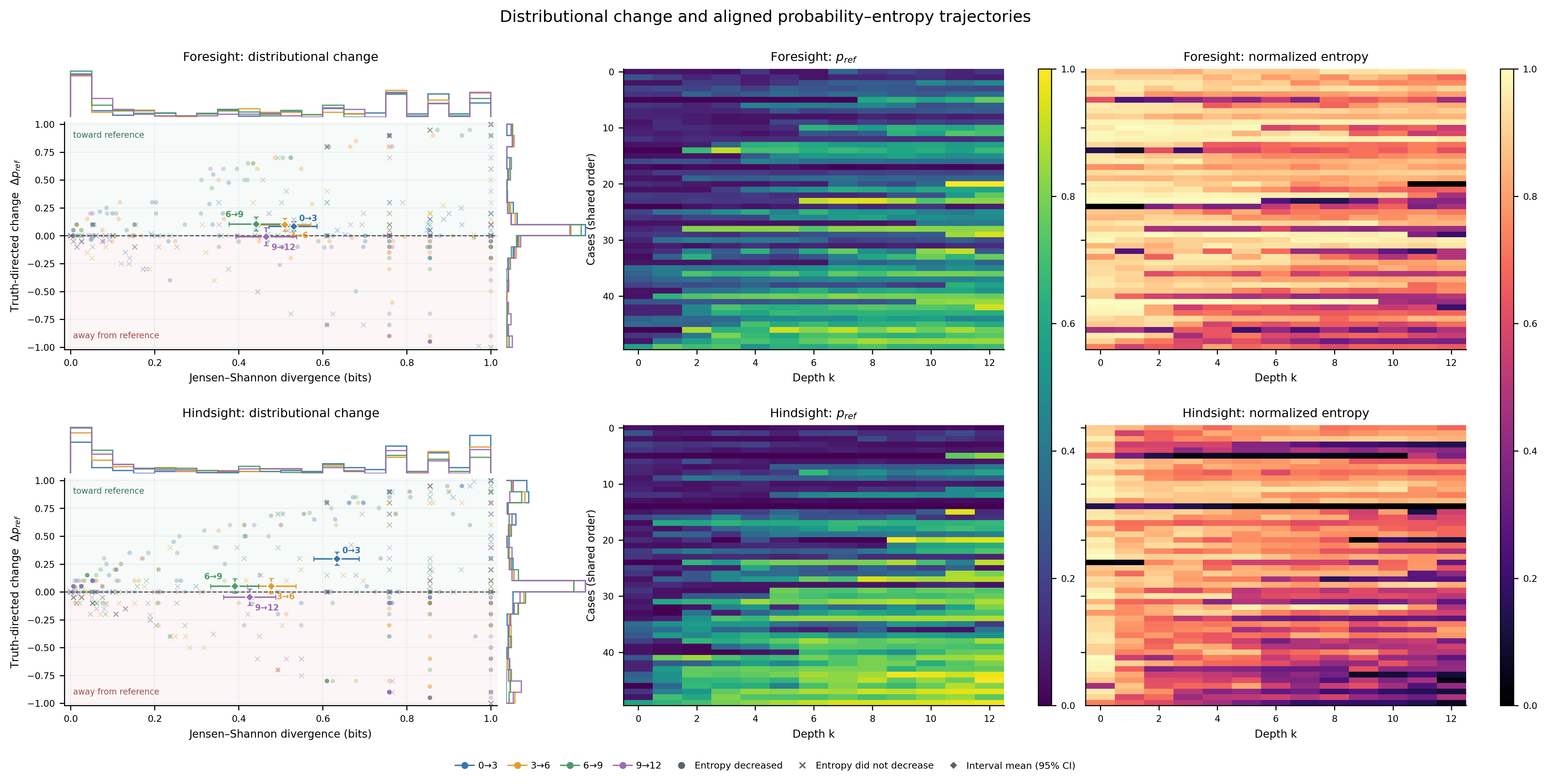}
\caption{Distributional change and diagnostic value.
Rows compare Foresight and Hindsight; columns show case-level JSD versus
$\Delta p_{\mathrm{ref}}$, aligned reference-probability trajectories, and
normalized-entropy trajectories.}
\label{fig:value-information}
\end{figure*}

\subsubsection{C.2 Distributional change $\neq$ diagnostic value}
\label{sec:exp-c3}

Figure~\ref{fig:value-information} provides two views of the same diagnostic
process. The top row shows Foresight and the bottom row Hindsight. In each
row, the left panel plots the magnitude of change between adjacent
diagnostic distributions (JSD) against its truth-directed component,
$\Delta p_{\mathrm{ref}}$; the middle and right panels show
$p_{\mathrm{ref}}$ and normalized entropy across all evidence depths for
the same ordered set of 50 cases. The scatter analysis includes only
intervals containing at least one newly acquired unique fact.

A large distributional change is not necessarily diagnostically useful:
at similar JSD values, $\Delta p_{\mathrm{ref}}$ may be positive, negligible,
or negative. This distinction is clearest at $0\!\rightarrow\!3$.
Hindsight produces only a moderately larger mean JSD than Foresight
(0.633 versus 0.531), but a much larger increase in reference-diagnosis
probability (0.298 versus 0.084). The heatmaps show the corresponding
state trajectories: Hindsight tends to increase $p_{\mathrm{ref}}$ earlier,
whereas entropy does not consistently track whether the diagnosis moves
toward the reference answer.

The figure therefore identifies two post-arrival failures. New evidence
may produce little diagnostic response, or it may substantially change the
distribution in the wrong direction. Evidence value consequently cannot
be inferred from update magnitude or concentration alone; it depends on
whether the update is truth-directed. These probability distributions are
elicited readouts rather than direct measurements of latent belief.

%% file: tables/results/C/evidence-value-decomposition.tex
\begin{table*}[t]
\centering
\caption{\textbf{Evidence-value decomposition.}
Panel A matches the realized evidence count case by case; Panel B reports
evidence identity and diagnostic stability through $k=12$ on the primary
cohort; Panel C reports paired set and order interventions.
Stable denotes a correct diagnosis at every remaining depth through $k=12$;
when fewer than $k$ non-initial facts are available, CER is computed over
the available Hindsight prefix. Gap and effect columns are percentage points.}
\label{tab:evidence-value-decomposition}

\begingroup
\fontsize{5.8}{6.8}\selectfont
\setlength{\arrayrulewidth}{0.45pt}
\setlength{\tabcolsep}{0pt}

% One outer row: continuous top/bottom rules and central divider.
\begin{tabular}{
@{}
p{0.515\textwidth}
@{\hspace{6pt}\vrule width 0.45pt\hspace{6pt}}
p{0.455\textwidth}
@{}
}
\hline

% ============================================================
% LEFT
% ============================================================
\begin{minipage}[t]{\linewidth}
\vspace{-2.5pt}

\begingroup
\setlength{\tabcolsep}{1.0pt}
\renewcommand{\arraystretch}{1.10}

% Panel A
\begin{tabularx}{\linewidth}{
@{}
>{\hsize=1.35\hsize\linewidth=\hsize\raggedright\arraybackslash}X
>{\hsize=.85\hsize\linewidth=\hsize\centering\arraybackslash}X
>{\hsize=.90\hsize\linewidth=\hsize\centering\arraybackslash}X
>{\hsize=1.10\hsize\linewidth=\hsize\centering\arraybackslash}X
>{\hsize=.65\hsize\linewidth=\hsize\centering\arraybackslash}X
>{\hsize=1.15\hsize\linewidth=\hsize\centering\arraybackslash}X
@{}
}
\rowcolor{tablegroup}
\multicolumn{6}{c}{
\textbf{Panel A: same number of evidence items}
}
\\
\noalign{\vskip 4pt}
\cmidrule(lr){1-6}

\shortstack[l]{\textbf{Cohort /}\\$\boldsymbol{k}$}
&
\shortstack{\textbf{Mean}\\$\boldsymbol{m_i(k)}$}
&
\textbf{Foresight}
&
\shortstack{\textbf{Matched}\\\textbf{Hindsight}}
&
\textbf{Gap}
&
\shortstack{\textbf{95\%}\\\textbf{CI}}
\\
\midrule

Primary / 3
& 2.09 & 33.1\% & 63.0\% & 29.8 & [22.1, 37.6]
\\
Primary / 6
& 3.52 & 42.0\% & 70.7\% & 28.7 & [20.4, 37.0]
\\
Continuation / 6
& 3.70 & 37.1\% & 72.9\% & 35.7 & [27.1, 44.3]
\\
Continuation / 12
& 5.96 & 47.9\% & 69.3\% & 21.4 & [12.1, 30.7]
\\
\end{tabularx}

\vspace{1.5pt}

% Panel C
\begin{tabularx}{\linewidth}{
@{}
>{\hsize=1.85\hsize\linewidth=\hsize\raggedright\arraybackslash}X
>{\hsize=.70\hsize\linewidth=\hsize\centering\arraybackslash}X
>{\hsize=.70\hsize\linewidth=\hsize\centering\arraybackslash}X
>{\hsize=.70\hsize\linewidth=\hsize\centering\arraybackslash}X
>{\hsize=1.25\hsize\linewidth=\hsize\centering\arraybackslash}X
>{\hsize=.80\hsize\linewidth=\hsize\centering\arraybackslash}X
@{}
}
\rowcolor{tablegroup}
\multicolumn{6}{c}{
\textbf{Panel C: direct set and order interventions ($N=100$)}
}
\\
\cmidrule(lr){1-6}

\textbf{Contrast}
&
\textbf{Ref.}
&
\textbf{Alt.}
&
\textbf{Effect}
&
\textbf{95\% CI}
&
\textbf{Measure}
\\
\midrule

Same-set reranking
& 34.3\% & 43.7\% & 9.3 & [5.5, 13.2] & AUSC
\\
Set replacement
& 43.7\% & 70.3\% & 26.7 & [19.0, 34.5] & AUSC
\\
Hindsight-order
& 44.3\% & 70.3\% & 26.0 & [20.2, 32.0] & AUSC
\\
\end{tabularx}

\endgroup
\vspace{1.5pt}
\end{minipage}

&

% ============================================================
% RIGHT
% ============================================================
\begin{minipage}[t]{\linewidth}
\vspace{-2.5pt}

\begingroup
\setlength{\tabcolsep}{1.0pt}
\renewcommand{\arraystretch}{1.03}

\begin{tabularx}{\linewidth}{
@{}
>{\hsize=.45\hsize\linewidth=\hsize\centering\arraybackslash}X
>{\hsize=.75\hsize\linewidth=\hsize\centering\arraybackslash}X
>{\hsize=1.50\hsize\linewidth=\hsize\centering\arraybackslash}X
>{\hsize=1.15\hsize\linewidth=\hsize\centering\arraybackslash}X
>{\hsize=1.15\hsize\linewidth=\hsize\centering\arraybackslash}X
>{\hsize=1.00\hsize\linewidth=\hsize\centering\arraybackslash}X
@{}
}
\rowcolor{tablegroup}
\multicolumn{6}{c}{
\textbf{Panel B: evidence identity and stable sufficiency ($N=181$)}
}
\\
\noalign{\vskip 4pt}
\cmidrule(lr){1-6}

$\boldsymbol{k}$
&
\textbf{CER}
&
\shortstack{\textbf{Hindsight}\\\textbf{Top-1 reached}}
&
\shortstack{\textbf{Foresight}\\\textbf{stable}}
&
\shortstack{\textbf{Hindsight}\\\textbf{stable}}
&
\shortstack{\textbf{Stable}\\\textbf{gap}}
\\
\midrule

1  & 5.5\%  & 5.5\%  & 12.7\% & 40.3\% & 27.6 \\
2  & 12.2\% & 13.3\% & 18.8\% & 45.3\% & 26.5 \\
3  & 16.0\% & 18.2\% & 20.4\% & 50.3\% & 29.8 \\
4  & 20.9\% & 23.8\% & 23.2\% & 50.8\% & 27.6 \\
5  & 23.8\% & 30.4\% & 25.4\% & 55.2\% & 29.8 \\
6  & 26.9\% & 35.4\% & 27.6\% & 58.6\% & 30.9 \\
7  & 28.6\% & 38.7\% & 34.3\% & 60.8\% & 26.5 \\
8  & 30.1\% & 42.5\% & 36.5\% & 62.4\% & 26.0 \\
9  & 31.6\% & 44.8\% & 39.2\% & 66.3\% & 27.1 \\
10 & 33.6\% & 45.9\% & 44.2\% & 68.0\% & 23.8 \\
11 & 34.6\% & 49.2\% & 49.2\% & 71.3\% & 22.1 \\
12 & 35.7\% & 53.6\% & 52.5\% & 77.3\% & 24.9 \\
\end{tabularx}

\endgroup
\vspace{1.5pt}
\end{minipage}

\\
\hline
\end{tabular}

\endgroup
\end{table*}

%% file: sections/results/D-after-evidence.tex
\begin{figure*}[t]
    \centering
    \includegraphics[width=\textwidth]{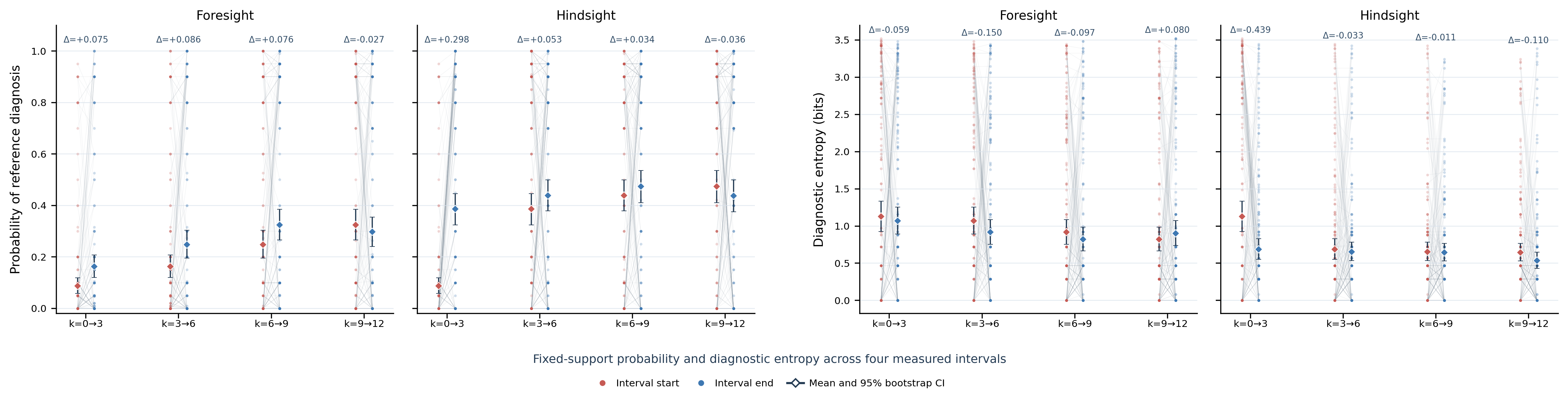}
    \caption{Diagnostic updating after evidence arrival from initial to round 12.}
    \label{fig:diagnostic-updating}
\end{figure*}

\subsection{D. After Evidence: Diagnostic Updating}
\label{sec:results-updating}

Section~\ref{sec:results-evidence} showed that foresight often acquires evidence with lower diagnostic value. We now ask what happens after evidence enters the case state that whether the diagnostic distribution changes, and whether a missed high-value fact can reverse an incorrect diagnosis.

\subsubsection{D.1 Diagnostic-state updating}
\label{sec:exp-d1}

Figure~\ref{fig:diagnostic-updating} follows the same 181 cases across four three-step intervals. Under hindsight, most updating occurs in the first interval which is from $k=0$ to $k=3$, the mean probability assigned to the reference diagnosis increases by 0.298, while diagnostic entropy changes by $-0.439$ bits. Later hindsight intervals show little additional change. Foresight increases the reference-diagnosis probability more gradually through $k=9$, but none of its entropy changes is clearly different from zero; from $k=9$ to $k=12$, it shows no reliable additional gain.
Thus, retrospectively prioritized evidence produces an early, concentrated diagnostic update, whereas prospectively acquired evidence yields slower and less consistent concentration. Together, these fixed-support probability distributions and entropy measures provide a behavioral proxy for how the model’s implicit diagnostic state changes as evidence accumulates.

\subsubsection{D.2 Controlled rescue by high-value evidence}
\label{sec:exp-d2}
The trajectory results alone cannot show whether a missed fact caused diagnostic failure. We therefore intervene on incorrect Foresight states by adding either an unseen hindsight ranked fact or a pre-frozen matched-control fact from the same case. In the prespecified complete-grid cohort, the mean critical-minus-control rescue is $+38.8$ percentage points for hindsight Rank 1, $+24.8$ for Rank 2, and $+10.3$ for Rank 3. The Rank-1-minus-Rank-3 contrast is $+28.5$ points. By contrast, the Rank-1 difference between insertion at $k=1$ and $k=12$ is not clearly supported.
These results show that omission of high-value evidence can directly contribute to diagnostic failure. Rescue depends more clearly on evidence rank than on the measured insertion depth. The all-eligible analysis is reported in the appendix: its critical-minus-control effect is positive in all 15 rank--depth cells, but its changing denominators do not permit a fixed-cohort comparison of insertion timing.

\begin{figure*}[t]
    \centering
    \includegraphics[width=\textwidth]{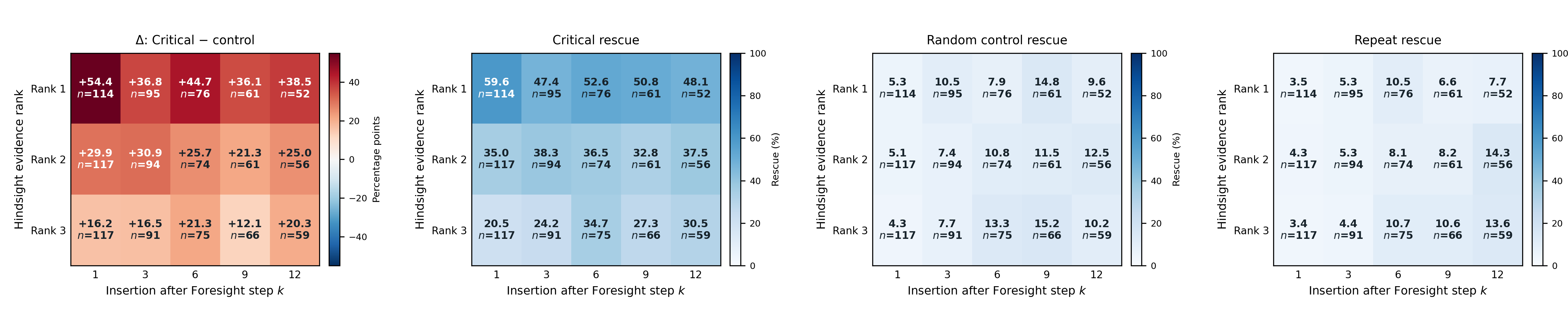}
    \caption{Diagnostic rescue by evidence rank and insertion depth.
Cells report rescue rates and eligible counts after inserting evidence. The left panel shows the critical-minus-random difference in percentage points, and the right three are individuals.}
    \label{fig:rank-timing-rescue}
\end{figure*}

%% file: sections/results/E-robustness-transfer.tex
\subsection{E. Robustness and External Validation}
\label{sec:results-transfer}

\begin{figure}[t]
    \centering
    \includegraphics[width=\linewidth]
    {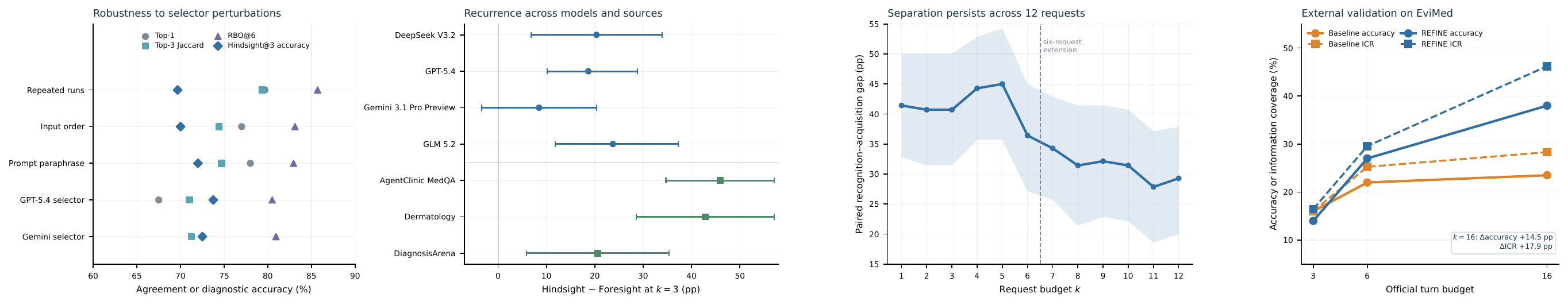}
    \caption{\textbf{Robustness and external validation.}
    The recognition--acquisition gap recurs across selector perturbations,
    model backbones, source cohorts, and longer horizons.}
    \label{fig:e-robustness}
\end{figure}

This gap recurs across implementation choices and
evaluation settings (Figure~\ref{fig:e-robustness}). On the shared 59-case
cohort, all four model backbones show positive paired gaps, with three
confidence intervals excluding zero; the gap is also positive within each
source cohort. In the separate 140-case continuation cohort, it remains
29.3 percentage points at $k=12$. Thus, the observed
separation is not attributable to a single selector, model, data source, or
short disclosure horizon.
External validation provides the same practical conclusion. On 200 balanced
EviMed cases, REFINE improves information coverage by 17.9 points and
diagnostic accuracy by 14.5 points over the official free-form baseline at
$k=16$, showing that stronger prospective acquisition can improve both
realized evidence and diagnosis. Full selector perturbations, model- and
source-specific estimates, EviMed comparisons, and controlled ablations are
reported in appendix. 

%% file: sections/appendix.tex
\appendix
% Keep supplementary tables at the surrounding font size unless they exceed
% the available line width. This avoids enlarging naturally narrow tables.
\newsavebox{\appendixtablebox}
\newcommand{\fitappendixtable}[1]{%
  \sbox{\appendixtablebox}{\input{#1}}%
  \ifdim\wd\appendixtablebox>\linewidth
    \resizebox{\linewidth}{!}{\usebox{\appendixtablebox}}%
  \else
    \usebox{\appendixtablebox}%
  \fi}
% Two compact, visually distinct frames for clinical examples and prompts.
% Both are defined here so the appendix remains self-contained in the existing
% manuscript without a new preamble dependency.
\newsavebox{\appendixframedcontent}
\newenvironment{clinicalcase}[1]{%
  \par\medskip\begingroup
  \setlength{\fboxsep}{7pt}\setlength{\fboxrule}{0.6pt}%
  \begin{lrbox}{\appendixframedcontent}%
  \begin{minipage}{\dimexpr\linewidth-2\fboxsep-2\fboxrule\relax}%
  \raggedright\small
  {\color{green!45!black}\bfseries Clinical case study: #1}\par\smallskip
}{%
  \end{minipage}\end{lrbox}%
  \noindent\fcolorbox{green!45!black}{green!3}{\usebox{\appendixframedcontent}}%
  \par\medskip\endgroup
}
\newenvironment{promptbox}[1]{%
  \par\medskip\begingroup
  \setlength{\fboxsep}{7pt}\setlength{\fboxrule}{0.6pt}%
  \begin{lrbox}{\appendixframedcontent}%
  \begin{minipage}{\dimexpr\linewidth-2\fboxsep-2\fboxrule\relax}%
  \raggedright\small
  {\color{blue!55!black}\bfseries Prompt: #1}\par\smallskip
}{%
  \end{minipage}\end{lrbox}%
  \noindent\fcolorbox{blue!55!black}{blue!2}{\usebox{\appendixframedcontent}}%
  \par\medskip\endgroup
}
\input{sections/appendix/00-methods-metrics.tex}
\input{sections/appendix/01-A-B.tex}
\input{sections/appendix/02-C.tex}
\input{sections/appendix/03-D.tex}
\input{sections/appendix/04-E.tex}
\input{sections/appendix/05-reproducibility.tex}

%% file: sections/appendix/00-methods-metrics.tex
\begingroup\sloppy
\section{Methods, Frozen Protocols, and Metric Definitions}
\label{app:methods-metrics}

\subsection{Study unit, evidence state, and interpretation}
\label{app:study-unit}

The unit of the controlled study is a case, not an API request. For case $c$, let
$x_c$ be its initial presentation, $y_c^*$ its reference diagnosis, and
$\mathcal U_c$ the realized non-initial atomic-fact inventory, with a stable
canonical ID for every fact. Let $E_c^q(k)$ be the \emph{set of unique fact IDs}
actually available after $k$ opportunities under condition $q$, and let
$L_c^q(k)$ be the corresponding \emph{ordered} list. A failed or duplicate
request advances $k$ but leaves $E_c^q(k)$ unchanged. A fresh diagnostic
probe reads $x_c$ and $L_c^q(k)$; its top diagnosis is $\hat y_c^q(k)$.
The initial state $k=0$ is shared. All main comparisons use the same cases and
the same independent semantic diagnosis rule.

\emph{Fullsight} sees the completed case at once and is used to establish
complete-case competence. \emph{Atomic Foresight} chooses one prospective
request at a time. \emph{Diagnosis-Unconditioned Hindsight} (Hindsight)
ranks facts after seeing $\mathcal U_c$, but without the reference diagnosis or
the Fullsight answer. The Hindsight prefix is a retrospective recognition
reference, not an executable acquisition policy. Withholding an explicit
diagnostic label does not remove the information advantage of seeing the
realized evidence. Case-level and source-level estimates therefore describe
the stated conditional cohorts, not unconditional clinical accuracy.

\subsection{Case flow and frozen analysis panels}
\label{app:cohorts}

The source universe was frozen before the final primary outcomes. It contains
320 cases: 140 AgentClinic MedQA, 80 dermatology, and 100 DiagnosisArena.
Fullsight was semantically correct on 190. Six of these were removed by the
recorded Fullsight quality-control exclusion and three dermatology cases had
fewer than six non-initial atomic facts. The remaining 181 cases were
protocol-complete, Fullsight-correct, and structurally eligible: 98
AgentClinic MedQA, 49 dermatology, and 34 DiagnosisArena. All six main action
slots and paired diagnostic prefixes are represented for these cases. The
exclusions were fixed before the paired analysis.

\begin{table}[ht]
\centering\small
\caption{Frozen study panels. Subsets overlap; their counts must not be added. A panel's own eligibility and denominator govern its result.}
\label{tab:app-cohort-flow}
\begin{tabular}{p{0.38\linewidth}rp{0.32\linewidth}}
\toprule
Panel & Cases & Purpose\\
\midrule
Source universe / Fullsight-correct / strict primary & 320 / 190 / 181 & Conditional six-step paired analysis\\
Result-blind menu & 120 & Frozen primary subset\\
Evidence-set/order intervention & 100 & Fixed six-prefix panel\\
Selector perturbation & 100 & Repeats, prompt, and order; alternative-model selector: 80\\
Four-model paired replication & 59 & Same case IDs across backbones\\
Shared four-model menu & 50 & Free/menu/Hindsight at three requests\\
Twelve-step continuation & 140 & Original policies continued for six slots\\
Fixed-support updating & 181 & Depths 0, 3, 6, 9, and 12\\
Rank-by-time rescue / complete grid & 140 / 33 & Eligible cells / fixed panel in every cell\\
EviMed external bridge & 200 & 50 each from four source families\\
\bottomrule
\end{tabular}
\end{table}

The earlier 12-model-slot screen yielded ten completed general-purpose
models and 798 eligible model--case records. The two medical-model slots
failed infrastructure gates and are not partial efficacy estimates. That
screen's retrospective condition and screening/retest calls differ from the
strict primary protocol; the joined records contain no identical screening
and reported request. We use it for breadth only, never pool its numerators
or denominators with the 181-case paired analysis.

For the external bridge, the distributed EviMed source pool contained 800
cases (200 from each of AgentClinic MedQA, DiagnosisArena, RareArena/RDC, and
Derm/CRAFT-MD). The source-ID overlap comparison removed 317 exact
internal overlaps, leaving 483 eligible cases; a balanced 50-per-source
sample of 200 was frozen for analysis. Normalized-text hash and similarity
checks supplemented the exact-ID comparison. The resulting EviMed denominator
is distinct from the internal 181-case denominator.

\subsection{Model roles, prompts, visibility, and deployment}
\label{app:roles-prompts}

The strict primary acquisition, ranking, and diagnostic model was
\texttt{deepseek/deepseek-v3.2} through OpenRouter. Four-model recurrence also
used \texttt{openai/gpt-5.4},
\texttt{google/gemini-3.1-pro-preview}, and \texttt{z-ai/glm-5.2} as the
condition's acquisition/diagnostic backbone on the shared cases. GPT-5.5 was
the independent diagnosis-equivalence grader. Selector, acquisition,
matcher, diagnostic probe, and grader are separate roles; none inherits
another role's unseen fields. The prompt wording, output schemas, parser
rules, and stage-specific settings are specified below.

\paragraph{Retrospective selector.}
The primary DeepSeek selector receives $x_c$ followed by the complete
non-initial pool, one line per \texttt{[FACT\_ID] fact text}, in canonical
order. Its instruction requires six distinct supplied IDs and JSON arrays
\texttt{ranked\_fact\_ids} and \texttt{rationales}, each of length six.
Exact-ID membership, uniqueness, and length are validated; no fuzzy remapping
is allowed. The Top-3 prefix is the principal
three-step comparator. Two frozen evidence-list permutations, repeated calls,
a prompt paraphrase, and GPT-5.5/Gemini selectors probe robustness. The
shared-model replication requested the frozen three-fact ranking for the
other backbones. These variants never change the primary selector after
observing outcomes.

\begin{promptbox}{Retrospective evidence selector}
\textbf{System.} Rank exactly six supplied non-initial atomic facts by
diagnostic usefulness for determining the diagnosis from the initial
presentation. You do not know the gold diagnosis, Fullsight diagnosis, or
Foresight trajectory. Use only supplied IDs, once each. Return JSON only:
\texttt{\detokenize{{"ranked_fact_ids":["..."],"rationales":["..."]}}}.
Both arrays must have length six.

\textbf{User template.} INITIAL PRESENTATION: \texttt{\{initial\_presentation\}}.
COMPLETE NON-INITIAL ATOMIC EVIDENCE POOL: one
\texttt{[FACT\_ID] fact text} per line, in canonical order.
Return the requested JSON ranking.
\end{promptbox}

Selector agreement is measured per case before averaging: Top-1 agreement
is an exact equality indicator; Top-3 Jaccard is
$|D_{c,a}^{(3)}\cap D_{c,b}^{(3)}|/
 |D_{c,a}^{(3)}\cup D_{c,b}^{(3)}|$. Rank-biased overlap at depth six uses
$a_d=|D_{c,a}^{(d)}\cap D_{c,b}^{(d)}|/d$ and
$\mathrm{RBO}@6=0.1\sum_{d=1}^{6}a_d(0.9)^{d-1}
 +a_6(0.9)^6$. Repeatability averages the three pairs from each of the
three frozen repeated rankings per case. Downstream robustness is a new
diagnostic probe on each alternative Top-3 prefix, with the same accuracy
rule and the variant's own case denominator.

\paragraph{Prospective action.}
At step $t$, the action prompt provides the initial presentation, previously
returned fact IDs and text, every earlier request and its status, the current
step, and remaining opportunities. Its system instruction states that the
policy cannot see hidden evidence, candidate IDs, a complete case, or any
diagnosis label. It asks for exactly one clinical information request and
JSON fields \texttt{request}, \texttt{request\_type} (history, physical exam,
lab, imaging, medication, or other), and \texttt{reason\_short}. A blank
request is invalid; an unrecognized request type is deterministically mapped
to \texttt{other}. The action model does not see the matcher's candidate
scores or the future result. The system and user templates are shown below.

\begin{promptbox}{Prospective information request}
\textbf{System.} You are the clinical acquisition policy. Use only the
supplied initial presentation, previously acquired facts, and prior request
outcomes. You cannot see hidden evidence, candidate IDs, a full case, or any
diagnosis label. Choose exactly one clinical information request for this step.
Return JSON only:
\texttt{\{}\texttt{\detokenize{"request":"free-text request",}}\allowbreak
\texttt{\detokenize{"request_type":"history|physical_exam|lab|imaging|medication|other",}}\allowbreak
\texttt{\detokenize{"reason_short":"brief rationale"}}\texttt{\}}.

\textbf{User template.} INITIAL PRESENTATION: \texttt{\{initial\_presentation\}}.
PREVIOUSLY ACQUIRED FACTS: \texttt{[FACT\_ID] fact text}, one per line.
PRIOR REQUEST OUTCOMES: for each earlier step, include its request, status
(matched, duplicate, or no-match), and returned fact or none.
CURRENT STEP $t$ OF 6; REMAINING OPPORTUNITIES INCLUDING THIS ONE:
\texttt{\{remaining\}}. Return the JSON action now.
\end{promptbox}

\paragraph{Deterministic realization.}
The frozen \texttt{m1\_lexical\_type\_v1} matcher sees the just-issued
request, its request type, $\mathcal U_c$, and acquired IDs. It tokenizes
case-folded alphanumeric words and Chinese-character bigrams, scores each
compatible fact by lexical cosine plus a $0.08$ type/modality bonus and a
$0.12$ exact-normalized-text bonus (capped at 1), and retains candidates with
score at least $0.10$ and at least $60\%$ of the top score. Canonical fact ID
breaks ties. The first selected ID is either new or duplicate; no candidate
is a no-match. These three statuses exhaust action outcomes and consume one
slot each. A request can release at most one new fact. No provider call,
manual override, or post-hoc clinical repair is made by the matcher.

\paragraph{Result-blind menu.}
The menu condition replaces free-text target generation with 103 global
clinical targets, frozen before case outcomes and shuffled deterministically
for each case and round. The action prompt shows the initial presentation,
observed acquired evidence, and target \emph{names}; it never shows their
unrealized results. It asks for one \texttt{selected\_target} ID in JSON.
The same one-fact, duplicate, and no-match accounting applies. A structured
target space changes the interface but does not supply hidden evidence or
equalize the facts ultimately obtained.

\begin{promptbox}{Result-blind evidence menu}
\textbf{System.} You are selecting the single next clinical evidence target
prospectively. Choose exactly one label from the global result-blind menu.
You see target names but never hidden results. Do not diagnose and do not
invent a label. Return JSON only as
\texttt{\detokenize{{"selected_target":"TARGET_ID","reason_short":"brief"}}}.

\textbf{User template.} INITIAL PRESENTATION: \texttt{\{initial\_presentation\}}.
OBSERVED ACQUIRED EVIDENCE: \texttt{[FACT\_ID] fact text}, one per line, or
``no acquired evidence yet.'' GLOBAL TARGET MENU:
\texttt{[TARGET\_ID] target name}, one per line in the case-round permutation.
\end{promptbox}

\paragraph{Fresh diagnostic probe and blinded grader.}
The diagnostic call is stateless with respect to prior diagnostic answers:
it receives only $x_c$ and the ordered available evidence prefix for the
current condition. Its system prompt requests JSON
\texttt{top\_diagnosis} and one to five unique ordered differential entries
with confidence in $[0,1]$, with no condition label, reference diagnosis,
future evidence, or multiple-choice options. A valid top diagnosis is moved
to the first differential position if necessary; percentage confidences in
$(1,100]$ are divided by 100. Exact normalized diagnosis matches are accepted
first. Otherwise an independent GPT-5.4 grader receives an opaque task ID,
the reference diagnosis, and the prediction only. It accepts standard
synonyms and rejects materially broader, narrower, or different diseases.
The grader does not see the acquisition condition or aggregate result. The
Final-v1 JSON output is \texttt{correct} plus a brief reason; unresolved
invalid outputs are not silently imputed. The prompts are shown below.

\begin{promptbox}{Fresh diagnostic probe}
\textbf{System.} You are a diagnostic reasoning probe. Use only the initial
presentation and available evidence in this request. Do not assume hidden
facts, future evidence, condition names, or multiple-choice options.
Return JSON only:\par
\noindent\texttt{\{"top\_diagnosis":"string",\allowbreak
"differential":[\{"diagnosis":"string",\allowbreak
"confidence":0.0\}]\}}\par
Provide 1--5 unique diagnoses, ordered most to least likely; the first must
equal \texttt{top\_diagnosis} and confidences must be in \([0,1]\).

\textbf{User template.} INITIAL PRESENTATION: \texttt{\{initial\_presentation\}}.
AVAILABLE CLINICAL EVIDENCE: each available fact as a separate bullet in the
condition-specific order. Return the requested JSON object now.
\end{promptbox}

\begin{promptbox}{Independent diagnosis-equivalence grader}
\textbf{System.} You are an independent semantic diagnosis equivalence grader.
Decide whether each candidate diagnosis is clinically equivalent to the
reference diagnosis, allowing standard synonyms but not broader, narrower,
or different diseases. Return JSON only as
\texttt{\detokenize{{"judgments":[{"task_id":"...","correct":true,"reason":"brief"}]}}}.

\textbf{User template.} For each item: TASK \texttt{\{opaque\_task\_id\}};
REFERENCE: \texttt{\{reference\_diagnosis\}}; CANDIDATE:
\texttt{\{candidate\_prediction\}}.
\end{promptbox}

\paragraph{Request settings and reproducibility.}
The Final-v1 protocol froze temperature 0 on 2026-09-17 and left $top_p$ at
the provider default; a provider-side deterministic seed was unavailable.
The primary selector used a 5,500-token output ceiling, raised to 6,500 only
for a frozen-prompt invalid/truncated-response recovery. Action calls used
1,600 tokens, diagnostic probes 3,500, and Final-v1 grader batches 6,000.
Invalid action/probe schemas could be retried up to three attempts under
the frozen prompt and schema reminder; selector recovery was a schema-only
retry under unchanged visibility. Failed attempts were retained in the
request ledger, and unresolved records were excluded until a valid rerun
without redefining cohort membership. The separate fixed-support extension
used JSON-object responses, low reasoning effort, a 7,500-token ceiling,
and its own missing-task recovery rule. The analysis environment used macOS
26.5.1 arm64 and Python 3.9.6. A provider-side deterministic seed was
unavailable, so API responses are not claimed to be byte-for-byte
reproducible.

\paragraph{Twelve-step continuation.}
The separate 140-case extension continued the \emph{existing} six-step
Atomic policy through slots 7--12; it did not restart a newly optimized
twelve-step policy. The first six action records and Hindsight ranking positions
were fixed. The Hindsight selector was prompted to append six distinct IDs from
the remaining realized pool, without reference diagnosis, Fullsight answer,
Atomic trajectory, or diagnostic outcomes. The Atomic prompt retained the
same system instruction and added an explicit extension-phase step/budget
line. Fresh diagnostic probes scored the continued prefixes. DeepSeek V3.2,
temperature 0, provider-default $top_p$, low reasoning effort, and the same
deterministic matcher were frozen for the formal run; stage-specific output
limits remained 5,500/6,500 selector, 1,600 action, 3,500 probe, 7,500
readout, and 6,000 grader tokens. A two-case pilot lacking the reasoning
setting was quarantined before expansion, and two billed JSON responses
were recovered from a provider reasoning field without new clinical calls.

\subsection{Primary diagnostic and evidence metrics}
\label{app:primary-metrics}

Let $\mathcal C$ be the relevant frozen case set, $N=|\mathcal C|$, and
$z_c^q(k)=\mathbf1[\hat y_c^q(k)\equiv y_c^*]$, where $\equiv$ is the
frozen semantic-equivalence judgment. Means give every case equal weight.
The primary confirmatory estimand is the paired three-step difference on the
strict 181-case set:
\begin{align}
 A_q(k)&=N^{-1}\sum_{c\in\mathcal C}z_c^q(k),\nonumber\\
 \Delta_{\rm Hindsight-F}(k)&=N^{-1}\sum_c
       \bigl(z_c^{\rm Hindsight}(k)-z_c^{\rm F}(k)\bigr),\label{eq:app-accuracy}\\
 \mathrm{AUSC}_q&=\frac{1}{6N}\sum_c\sum_{k=1}^{6}z_c^q(k),\nonumber\\
 \Delta_{\rm AUSC}&=\mathrm{AUSC}_{\rm Hindsight}-\mathrm{AUSC}_{\rm F}.
 \label{eq:app-ausc}
\end{align}
The average under the six discrete prefixes is an arithmetic mean, not a
continuous-time area or a per-request success probability. Other $k$ values,
extensions, and perturbation panels are descriptive unless separately
specified. At $k=0$, the same initial diagnostic call is reused for paired
conditions.

For each case, let $n_c(k)=|E_c^{\rm F}(k)|$ and let $D_c^{(r)}$ be the first
$r$ unique canonical IDs in its Hindsight ranking. The action/evidence metrics are
\begin{align}
 \mathrm{Yield}(k)&=\frac1N\sum_c\frac{n_c(k)}{k},\qquad k>0,\nonumber\\
 \mathrm{CER}@k&=\frac1N\sum_c
       \frac{|E_c^{\rm F}(k)\cap D_c^{(k)}|}{k},\label{eq:app-cer}\\
 \mathrm{Top1Reach}(k)&=\frac1N\sum_c
       \mathbf1[D_c^{(1)}\subseteq E_c^{\rm F}(k)],\nonumber\\
 \mathrm{Top1Never}@6&=1-\mathrm{Top1Reach}(6).
 \label{eq:app-topone}
\end{align}
In the six-slot primary path, FillRate is $\sum_c n_c(6)/(6N)$;
no-match and duplicate rates use the corresponding number of statuses over
the same $6N$ slots. Thus FillRate plus the two failure rates equals one.
CER is exact-ID overlap with the \emph{first $k$} Hindsight facts; it is neither
semantic recall nor recall of a clinician-authored critical-fact gold set.
The order of facts is irrelevant to CER but relevant to diagnostic probes.

The casewise actual-fact-count control defines $m_c(k)=n_c(k)$ and compares
Foresight@${k}$ with a fresh diagnosis after the first $m_c(k)$ Hindsight-ranked
facts. If $m_c(k)=0$, both conditions use the initial state. Its accuracy gap
is the paired mean of $z_c^{\rm Hindsight}(m_c(k))-z_c^{\rm F}(k)$ over all 181
cases; no failed-fill case is dropped. Equal fact count does not imply equal
fact identity, order, or prospective visibility.

Define stable correctness for a six-prefix path as
\begin{align}
 k_c^{\mathrm{stable},q}
  &=\min\bigl\{j\in\{0,\ldots,6\}:z_c^q(t)=1\ 
       \forall t\in\{j,\ldots,6\}\bigr\},\nonumber\\
 S_q(k)&=N^{-1}\sum_c\mathbf1[k_c^{\mathrm{stable},q}\le k].
 \label{eq:app-stable}
\end{align}
If no such $j$ exists, $k_c^{\mathrm{stable},q}>6$. This differs from the
first correct prefix: a later relapse prevents an earlier prefix from being
called stable. Casewise trajectories are evaluated at every available prefix.

For the 100-case fixed order panel, $FF$ is the originally obtained Foresight
fact set in acquisition order, $FR$ is that same obtained set reranked by an
independent selector call, $HH$ is the Hindsight set in its selected order, and $HR$
is the Hindsight set in reverse order. For each condition the six-prefix AUSC is
computed as in Eq.~\eqref{eq:app-ausc}, with repeated unchanged evidence when
the Foresight set contains fewer than six facts. The three reported paired
contrasts are
\begin{align}
 G_{\mathrm{order,F}}&=\overline{\mathrm{AUSC}_{FR}-\mathrm{AUSC}_{FF}},\nonumber\\
 G_{\mathrm{set}}&=\overline{\mathrm{AUSC}_{HH}-\mathrm{AUSC}_{FR}},\nonumber\\
 S_{\mathrm{order,H}}&=\overline{\mathrm{AUSC}_{HH}-\mathrm{AUSC}_{HR}}.
 \label{eq:app-order}
\end{align}
Bars denote the mean over the fixed 100 cases. These are interventions on
presentation/set for an independently queried diagnostic model, not an
additive decomposition of a prospective policy's avoidable error.

\subsection{Action adaptation and diagnostic transitions}
\label{app:transition-metrics}

For the 181-case six-step trajectories, each adjacent pair $k\to k+1$ for
$k=1,\ldots,5$ is classified $WW,WC,CC,$ or $CW$ from diagnostic correctness.
The pooled transition denominator is the number of eligible \emph{pairs},
with case-cluster resampling for uncertainty. In particular,
\begin{align}
 \mathrm{WSP}&=\frac{\#WW}{\#WW+\#WC},\qquad
 P(W\!\to\!C\mid W)=\frac{\#WC}{\#WW+\#WC},\nonumber\\
 P(C\!\to\!W\mid C)&=\frac{\#CW}{\#CC+\#CW}.
 \label{eq:app-wsp}
\end{align}
Correct-state retention is $\#CC/(\#CC+\#CW)$. Evidence-conditioned WSP
uses the same wrong-origin denominator restricted to whether the \emph{next}
action returned a new fact. The always-wrong measure is the proportion of
cases wrong at every $k=1,\ldots,6$. These measures are observational state
transitions; they do not attribute all $W\to C$ changes to the immediately
preceding fact.

Action stagnation rate is the six-slot fraction of duplicate or no-match
outcomes; longest stagnation run counts consecutive such outcomes within a
case. A failure-recovery opportunity is a duplicate/no-match action at
$t\le5$; recovery is a new fact at $t+1$. Hence
\begin{equation}
 \mathrm{FRR}=\frac{\#\{t\le5:\mathrm{failure}_t,
                  \mathrm{new}_{t+1}\}}
 {\#\{t\le5:\mathrm{failure}_t\}}.
 \label{eq:app-frr}
\end{equation}
The request-type and target-modality adaptation rates have that same
eligible failure-follow-up denominator, with a changed type or modality in
the numerator. Request-text novelty is $1-$TF--IDF cosine similarity between
the failed request and its immediate successor, averaged within eligible
cases. Stratified recovery rates use the corresponding failure-type,
position, or switching subset as denominator; they describe associations,
not randomized strategy effects. At Hindsight rank cutoff $r\in\{1,3,6\}$,
define $\mathcal T_r$ as all $(c,k)$ with $k=1,\ldots,5$, an incorrect
Foresight diagnosis at $k$, and a newly returned fact at $k+1$ belonging
to $D_c^{(r)}$. Evidence-conditioned nonresponse (ECNR@$r$) is
\begin{equation}
 \mathrm{ECNR}@r=\frac{1}{|\mathcal T_r|}
  \sum_{(c,k)\in\mathcal T_r}\mathbf1[z_c^{\rm F}(k+1)=0].
 \label{eq:app-ecnr}
\end{equation}
Event and case denominators are specified for each analysis.

\subsection{Controlled one-fact rescue}
\label{app:rescue-metrics}

Two distinct frozen interventions test diagnostic response after an incorrect
state. In the earlier 114-case Top-1 rescue, eligibility was fixed before
new outcomes: primary membership, incorrect Atomic@3, absent Hindsight Top-1 in the
first three realized facts, a valid Hindsight Top-1, and a complete Atomic@3 probe.
$S0$ is the original wrong answer; $S1$ appends the missing Hindsight Top-1 to the
same prefix; $S2$ appends one preselected unseen control fact to that prefix.
The control excludes initial, acquired, and Hindsight Top-3 facts; same-modality
controls are preferred and a documented fallback is used otherwise. Frozen
SHA-256 ordering, prompt hashes, and complete evidence sequences were fixed
before all outcome calls. The $S1$ wrong-to-correct rate is $N_e^{-1}\sum_c z_c^{S1}$
because all $S0$ states are wrong, and the paired critical-minus-control
effect is $N_e^{-1}\sum_c(z_c^{S1}-z_c^{S2})$, with $N_e=114$.

The later rank-by-insertion-depth experiment evaluates Hindsight ranks 1--3 at
$k\in\{1,3,6,9,12\}$, always comparing immediate fresh-context correctness
after a critical fact with correctness after a pre-frozen matched control.
For cell $(r,k)$, both rescue rates and their difference use that cell's
eligible paired cases as denominator. The all-eligible denominators vary
across cells (91--117); a 33-case complete-grid panel supplies the fixed
denominator for rank and timing contrasts. Controls exclude all facts
acquired through action 12 and Hindsight ranks 1--6, with deterministic fallback and
SHA-256 choice. Neither experiment
estimates a deployable policy that could know the missing fact prospectively.

\begin{promptbox}{One-fact rescue diagnostic probe}
\textbf{System.} You are a diagnostic reasoning probe. Use only the initial
presentation and available evidence in this request. Do not assume hidden
facts, future evidence, condition names, or multiple-choice options. Return JSON only:\par
\noindent\texttt{\{"top\_diagnosis":"string",\allowbreak
"differential":[\{"diagnosis":"string",\allowbreak
"confidence":0.0\}]\}}.\par
Provide 1--5 unique diagnoses, ordered most to least likely; the first must
equal \texttt{top\_diagnosis} and confidences must be in $[0,1]$.

\textbf{User template.} INITIAL PRESENTATION: \texttt{\{initial\_presentation\}}.
AVAILABLE CLINICAL EVIDENCE: each fact in the fixed prefix on its own line,
followed by either the inserted Hindsight-ranked fact or the preselected control
fact. Return the requested JSON object now.
\end{promptbox}

\subsection{Standardized elicited distribution and value metrics}
\label{app:readout-metrics}

The probabilistic readout is a separate elicitation call, not the model's
internal belief. For each case, a fixed support of 8--12 diagnosis candidates
plus \texttt{OTHER} was constructed from pre-existing differential outputs
with semantic deduplication and the reference answer retained. The same
support is used across the compared evidence states. The model sees candidate
labels and available clinical evidence, but not the reference indicator. It
must assign nonnegative mass to every fixed candidate ID; raw sums must be
within $0.01$ of one, and invalid vectors are not arbitrarily normalized.
The support is fixed for each case before the compared readouts.

\begin{promptbox}{Fixed-support probability readout}
\textbf{System.} You are a clinical diagnostic probability allocator. Each
task is independent. Use only the clinical state shown for that task.
Allocate probabilities from 0 to 1 across exactly the supplied candidate
IDs including OTHER. The values for each task must sum to 1. Do not add,
omit, rename, or merge candidate IDs. omit, rename, or merge candidate IDs. Return JSON only as\par
\noindent\texttt{\{"readouts":[\{"task\_id":"...",\allowbreak
"probabilities":\{"candidate\_id\_01":0.0,\allowbreak
"OTHER":0.0\}\}]\}}.\par

\textbf{User template.} For each independent task, provide its opaque task
ID, initial presentation, currently available clinical evidence, and the
fixed list of candidate IDs with diagnosis labels. Assign mass to every
candidate ID and OTHER.
\end{promptbox}

For accepted vector $p_{c,k}^q$ and the support index $y_c^*$, define
$\widetilde p_*(p)=10^{-6}$ when $p_{y_c^*}=0$, and
$\widetilde p_*(p)=p_{y_c^*}$ otherwise. Then
\begin{align}
 H(p)&=-\sum_j p_j\log_2 p_j,\nonumber\\
 S_*(p)&=-\log_2\widetilde p_*(p),
 \label{eq:app-entropy}\\
 \mathrm{UR}(a,b)&=H(p_a)-H(p_b),\nonumber\\
 \mathrm{TDG}(a,b)&=S_*(p_a)-S_*(p_b)\nonumber\\
 &=\log_2\frac{\widetilde p_*(p_b)}{\widetilde p_*(p_a)}.
 \label{eq:app-tdg}
\end{align}
The $10^{-6}$ floor applies only when reference mass is exactly zero in the
frozen implementation; entropy and Brier scores retain the raw accepted
probabilities. A $10^{-9}$ floor sensitivity is reported because zero-mass
transitions materially affect mean TDG. Positive UR means concentration;
positive TDG means increased probability of the reference diagnosis. The
two signs need not agree. The trajectory summaries also report
$\Delta p_*=p_{b,y_c^*}-p_{a,y_c^*}$ and the paired mean entropy or
reference-probability change over $0\to3$, $3\to6$, $6\to9$, and $9\to12$.
The $9$ and $12$ states are the separately frozen 181-case extension, not
unreported points interpolated from the six-step primary readouts.

Jensen--Shannon divergence measures \emph{magnitude} of change:
\begin{equation}
 \mathrm{JSD}(p,q)=\tfrac12D_{\rm KL}(p\Vert m)
                  +\tfrac12D_{\rm KL}(q\Vert m),\quad m=(p+q)/2,
 \label{eq:app-jsd}
\end{equation}
with base-2 logarithms and zero-mass terms defined by continuity. The
distributional-change scatter excludes intervals with no newly acquired
unique fact; this restriction is a denominator condition, not a judgment
about diagnostic quality. Normalized entropy, when plotted, divides $H(p)$
by $\log_2|\mathrm{support}_c|$. The one-hot Brier score is
$\sum_j(p_j-\mathbf1[j=y_c^*])^2$; multiclass log loss in the
calibration table is $-\ln\max(p_{y_c^*},10^{-6})$; ten-bin ECE is
$\sum_b (n_b/n)|\bar p_b-\bar a_b|$, using the elicited top candidate's
confidence and its observed correctness. We do not interpret calibration,
JSD, or entropy as a direct latent-state measurement.

The descriptive quadrant labels use two frozen tolerances,
$\epsilon_{\rm UR}=0.02$ and $\epsilon_{\rm TDG}=0.05$ bits. Positive
UR/TDG is productive narrowing; negative UR/positive TDG is productive
broadening; positive UR/negative TDG is overconfident misdirection; both
negative is diffuse deterioration; observations within either tolerance are
neutral/mixed. Readout reliability is quantified by same-state Top-1
agreement, Spearman correlation of reference probability and entropy, and
median JSD under identical repeats, candidate-order permutations, prompt
paraphrase, and candidate-support replacement. The initial reliability
assessment sampled
40 initial, 40 Foresight@3, and 40 Hindsight@3 states. The reliability target set
was not fully met, so probabilistic results are sensitivity analyses.

For the frozen local counterfactual fact panel at a decision point, each
candidate $f$ is appended separately to the \emph{same} evidence prefix and
assigned realized $\mathrm{TDG}(f)$. The panel contains the actual selected
fact when matched, unseen Hindsight-ranked facts, and random unseen facts, with at
most six unique entries; it is not a global search. Among $m\ge2$ valid
candidates and one matched actual choice, average-rank ties give
\begin{align}
 \mathrm{CVP}&=\frac{\mathrm{rank}_{\uparrow}
                    (\mathrm{TDG}_{\rm chosen})-1}{m-1},&
 \mathrm{TDAR}&=\max_{f\in\mathrm{panel}}\mathrm{TDG}(f)
                    -\mathrm{TDG}_{\rm chosen}.\label{eq:app-value}
\end{align}
TopValueHit@1/@2 indicate whether the chosen fact has descending TDG rank
at most 1/2. The original panel has 200 decision points from 100 cases, of
which 138 points in 92 cases meet the actual-choice estimability rule. The
later free/menu/random policy comparison reports each policy's estimable
denominator separately; it does not assume that identical percentages imply
identical cases. TDAR is nonnegative by construction, and realized TDG is
not prospective expected information gain.

\subsection{External information coverage and statistical inference}
\label{app:stats}

For EviMed, $G_c$ denotes the official indexed reference-fact set and
$E_c^q(k)$ the unique reference IDs returned by the patient/reporter
simulator under strategy $q$. Information Coverage Rate is the \emph{mean
of case-level fractions},
\begin{equation}
 \mathrm{ICR}_q(k)=\frac1N\sum_{c\in\mathcal C}
     \frac{|E_c^q(k)\cap G_c|}{|G_c|}.
 \label{eq:app-icr}
\end{equation}
It is not pooled numerator divided by pooled denominator. The external run
used the EviMed role definitions and REFINE sequence, a 16-turn budget, and DeepSeek V3.2
doctor/patient/reporter/summarizer/diagnostician/verifier roles at
temperature 0. Its batched transport adapter multiplexed independent cases
under opaque IDs while retaining case-local histories and indexed simulator
facts. The menu uses the
same result-blind 103-target pool. The Hindsight row is a retrospective reference
with greater realized evidence yield, not an equal-fact prospective arm.

For the main paired diagnostic endpoint and other Final-v1 estimates, 95\%
intervals are percentile intervals from 5,000 case bootstrap resamples with
seed 20260917. In each draw the two conditions of a case remain paired. The
four-model panel samples shared case clusters so all four model records for
one case remain together. The exact $k=3$ McNemar analysis uses only
discordant paired diagnoses: if $b$ cases are Hindsight-only correct and $c$ are
Foresight-only correct, the matched-pair odds ratio is $b/c$ and the
two-sided exact $p$ value is from $\operatorname{Binomial}(b+c,1/2)$;
concordant pairs do not contribute. Bootstrap intervals and this exact
discordance test answer different questions. Six-step trajectory dynamics
and the 114-case rescue use 10,000 case-cluster draws with seed 20260920;
the later rank-by-depth rescue uses 10,000 draws with seed 20260924.
Fixed-support 12-step updates use their own 10,000-case bootstrap with seed
20260924. Every cell-specific estimate retains its own observed denominator;
no missing output is imputed inside the bootstrap.
\endgroup

%% file: sections/appendix/01-A-B.tex
\section{Experiment A: Capability contrast}
\label{app:ability}

\subsection{What the broad screen establishes}
\label{app:a1}

The opening ten-model comparison is a \emph{descriptive discovery screen}.
Twelve model slots were planned; ten general-purpose models completed all
required stages, whereas Ling 3.0 Flash Sante and MedGemma 27B IT failed
infrastructure gates. The ten complete rows contribute 798 eligible
model--case pairs, with model-specific denominators of 63--92. The
eligible-pair-weighted accuracies are 91.0\% for complete-case Fullsight,
74.2\% for prospective Foresight, and 86.5\% for the earlier retrospective
condition. Thus the descriptive Fullsight--Foresight difference is 16.8
percentage points (pp), partitioned arithmetically into 4.5 pp between
Fullsight and the early retrospective condition and 12.3 pp between that
condition and Foresight. Nine of ten complete models have a positive
retrospective-minus-prospective point estimate; GLM 5.2 ties. Grok 4.6's
retrospective point estimate exceeds its Fullsight estimate, illustrating
that these are noisy empirical accuracies, not an ordered mathematical bound.

\begin{table}[H]
\caption{All ten completed models in the early capability screen. Values are
percent correct among each model's eligible cases. The early retrospective
selector was diagnosis-conditioned and is not pooled with the strict
diagnosis-unconditioned experiment.}
\label{tab:app-a-broad}
\centering\small
\input{tables/appendix/new/broad-ten-models.tex}
\end{table}

The screen's retrospective ranking had access to a completed diagnosis and
its own completed-case fact inventory. Its model-specific eligibility and
earlier protocol differ from the primary DeepSeek V3.2 comparison below. The
Fullsight screen--retest audit also identifies separate calls: 0.0\% of
joined screening and reported records share a request ID, and 9.1\% change
correctness. Temperature zero therefore does not turn retesting into an
identical-request control. No medical-versus-general-model comparison is
made because neither planned medical model completed the infrastructure
gate.

\subsection{Strict cohort, visibility, and paired estimand}
\label{app:a2}

The frozen source universe has 320 cases: 140 AgentClinic MedQA, 80
dermatology, and 100 DiagnosisArena. Fullsight was semantically correct on
190/320 (59.4\%). Six of these 190 were excluded by recorded Fullsight
quality-control reasons and three lacked six non-initial canonical facts.
The primary cohort therefore contains 181 cases (98 AgentClinic MedQA, 49
dermatology, 34 DiagnosisArena). Every primary estimate conditions on
Fullsight correctness, protocol completeness, and at least six non-initial
facts; it is not unconditional clinical accuracy. Fullsight receives the
entire case. Atomic Foresight sees the initial presentation, its already
returned facts and previous request outcomes, and one further action is
allowed per step. The strict Hindsight selector sees the initial
presentation and complete \emph{non-initial} fact inventory, but no reference
diagnosis, Fullsight diagnosis, Foresight trajectory, or downstream
diagnostic outcome. A fresh diagnostic call receives only the initial
presentation and the ordered evidence prefix for its condition.

For case $c$, let $Y_c^F(k)$ and $Y_c^H(k)$ indicate semantic correctness of
the top diagnosis after $k$ prospective requests and $k$ retrospectively
ranked facts, respectively. The paired difference is
\begin{equation}
  \widehat\Delta(k)=\frac{1}{N}\sum_{c=1}^{N}
  \left[Y_c^H(k)-Y_c^F(k)\right].
  \label{eq:app-a-gap}
\end{equation}
An action with no match or a duplicate still consumes one Foresight budget
slot; the same evidence prefix may therefore be diagnosed at consecutive
values of $k$. Let $n_{01}$ denote cases wrong under Foresight and correct
under Hindsight and $n_{10}$ the reverse. The matched-pair odds ratio is
$n_{01}/n_{10}$, and the two-sided exact McNemar test conditions on
$n_{01}+n_{10}$ discordant cases. The 95\% intervals for
$\widehat\Delta(k)$ are percentile intervals from 5,000 paired case
bootstrap resamples, seed 20260917. The $k=3$ difference was the frozen
confirmatory endpoint; the other budgets are descriptive.

\begin{table}[H]
\caption{Primary paired accuracy at every original request budget
($N=181$). The interval is for the within-case Hindsight-minus-Foresight
difference, not for two independent accuracies.}
\label{tab:app-a-primary}
\centering\small
\input{tables/appendix/A/primary-paired.tex}
\end{table}

At $k=3$, 60/181 prospective and 133/181 retrospective diagnoses are
correct (33.1\% and 73.5\%). The 40.3-pp paired difference has a 95\%
interval of 32.6--48.6 pp. The 78 cases correct only under Hindsight and
five correct only under Foresight give an odds ratio of 15.60 and an exact
two-sided McNemar $p=6.40\times10^{-18}$. At $k=6$, accuracies are 42.0\%
and 75.7\%, leaving 33.7 pp (95\% CI 26.0--41.4). These facts support a
large separation between recognition after results are visible and
prospective acquisition. They do not quantify how much of the separation
could be removed by an implementable policy with only prospective
information.

\paragraph{Selector and deployment record.}
The primary selector was DeepSeek V3.2 via OpenRouter, frozen on 17
September 2026, temperature 0, provider-default top-$p$, nominal
\texttt{max\_tokens}=5,500 (6,500 only in a frozen-prompt response recovery).
Its system instruction was: \emph{``Rank exactly six supplied non-initial
atomic facts by diagnostic usefulness for determining the diagnosis from
the initial presentation. You do not know the gold diagnosis, Fullsight
diagnosis, or Foresight trajectory. Use only supplied IDs, once each.''}
It requested JSON arrays \texttt{ranked\_fact\_ids} and
\texttt{rationales}, each of length six. The user message supplied the
initial presentation and one \texttt{[FACT\_ID]}--text line per complete
non-initial fact in canonical order. Invalid, repeated, or out-of-pool IDs
were rejected; there was one schema-only retry, while transport recovery
reused the frozen prompt. Case and order permutations used seed 20260918;
the provider exposed no deterministic generation seed.

Atomic action generation and diagnostic probing were separate DeepSeek V3.2
calls. The action output limit was 1,600 tokens and the diagnostic output
limit 3,500; a separate GPT-5.5 semantic judge used up to 6,000 tokens.
All ran at temperature zero and provider-default top-$p$. The action
prompt required a single free-text request, one declared request category,
and a short reason while explicitly prohibiting hidden facts, candidate
IDs, the complete case, and diagnoses. The diagnostic probe requested one
top diagnosis and 1--5 ordered differential entries, without condition
name or multiple-choice options. The judge saw an opaque task ID,
reference diagnosis, and prediction but not experimental condition.
Temperature zero does not guarantee identical provider outputs.

\paragraph{Twelve-request continuation.}
The 140-case continuation is a frozen \emph{subset} of the 181 primary
cases, requiring at least 12 non-initial facts. It extends the original
six-step Foresight trajectory by six actions; it is not a new policy
optimized from the first turn for a 12-step horizon, nor an independent
sample. The original six hindsight IDs remain fixed, and the selector
chooses six additional distinct IDs from the remaining complete pool
without seeing diagnosis or earlier trajectories. The continuation used
the same OpenRouter DeepSeek V3.2 acquisition/diagnosis model and GPT-5.5
semantic judge, temperature zero, provider-default top-$p$, low reasoning
effort, and a frozen prompt for the extension budget. At $k=12$ in this
fixed subset, 67/140 Foresight and 108/140 Hindsight diagnoses are correct
(47.9\% versus 77.1\%; difference 29.3 pp, 95\% CI 20.0--37.9). The
181-case primary protocol ends at $k=6$; no 181-case $k>6$ value is
estimated. The extension demonstrates persistence over the observed
additional requests, not independent replication. Full round-by-round
estimates for the primary and continuation cohorts are in
Table~\ref{tab:primary}.

\begin{clinicalcase}{A.2: Missed injury mechanism and tendon function}
In AgentClinic case \texttt{agentclinic\_medqa\_14}, a 22-year-old man
initially reports right ring-finger pain and swelling for two days. The
reference condition is flexor digitorum profundus tendon avulsion
(``jersey finger''). In six prospective requests, the policy acquires
progressive swelling (\texttt{P5}), distal interphalangeal tenderness
(\texttt{E6}), a blood-pressure entry (\texttt{E2}), and a hand radiograph
describing soft-tissue swelling (\texttt{E12}); the third request repeats
\texttt{E6}, and the sixth returns no match. Its top diagnosis moves from
cellulitis to septic/infectious arthritis and remains incorrect at $k=6$.
The diagnosis-unconditioned retrospective selector ranks the jersey-pull
mechanism (\texttt{P4}), inability to flex the distal interphalangeal joint
(\texttt{E7}), and absence of fracture (\texttt{E11}) first. None of these
three was acquired prospectively. The fresh Hindsight diagnosis becomes
correct after \texttt{P4,E7} and stays correct through $k=6$. This trace
shows the central contrast at the level of clinical observations, while
the retrospective selector's access to realized results prevents it from
serving as a prospective clinical policy.
\end{clinicalcase}

\section{Experiment B: Before evidence arrives}
\label{app:before-evidence}

\subsection{A request is not a new observation}
\label{app:b1}

Let $a_{ct}$ be the one action attempted at turn $t$ and $m(a_{ct})$ the
deterministic match to the canonical non-initial fact inventory. A request
has outcome \emph{new} if the matcher returns a fact ID not previously
acquired, \emph{duplicate} if the selected ID is already acquired, and
\emph{no match} if no candidate passes the matcher. The three categories
are mutually exclusive, exhaustive, and each consumes a turn. For the
ordered set of unique acquired fact IDs $E_c(k)$ after $k$ requests, mean
realized fact count and per-slot evidence yield are
\begin{equation}
 \overline U(k)=N^{-1}\sum_c |E_c(k)|,\qquad
 F(k)=N^{-1}\sum_c |E_c(k)|/k=\overline U(k)/k.
 \label{eq:app-b-yield}
\end{equation}
These count observed canonical facts, not clinical usefulness or successful
reasoning. The primary cohort has exactly $181\times6=1{,}086$ request
slots: 637 new facts (58.7\%), 275 duplicates (25.3\%), and 174 no
matches (16.0\%). Thus $\overline U(3)=379/181=2.09$ and
$\overline U(6)=637/181=3.52$, despite nominal budgets of three and six.

\begin{table}[H]
\caption{Action realization by request position in the 181-case primary
cohort. Every position has 181 attempts. Cumulative unique facts count
newly returned IDs, so duplicate and no-match turns do not increase it.}
\label{tab:app-b-realization}
\centering\small
\begin{tabular}{rrrrrr}
\toprule
Turn & New & Duplicate & No match & New/181 & Mean unique by turn \\
\midrule
1 & 164 & 0 & 17 & 90.6\% & 0.91 \\
2 & 124 & 25 & 32 & 68.5\% & 1.59 \\
3 & 91 & 50 & 40 & 50.3\% & 2.09 \\
4 & 95 & 55 & 31 & 52.5\% & 2.62 \\
5 & 88 & 64 & 29 & 48.6\% & 3.10 \\
6 & 75 & 81 & 25 & 41.4\% & 3.52 \\
\midrule
All & 637 & 275 & 174 & 58.7\% & --- \\
\bottomrule
\end{tabular}
\end{table}

The frozen matcher \texttt{m1\_lexical\_type\_v1} sees the free-text
request, declared category, complete canonical non-initial fact texts, and
IDs already returned; the policy never sees its candidate list before
choosing. Candidate score is lexical cosine similarity plus a 0.08
category/modality bonus and a 0.12 exact normalized-text bonus, capped at
one. Candidates must score at least 0.10 and at least 60\% of the top
score; ties use canonical fact ID. The first candidate is returned, with
no model-based repair or manual override.
Matching is an operational simulator, not proof that a human interlocutor
would return the same fact. In particular, a lexical match can be clinically
weak or modality-incongruent.

\begin{clinicalcase}{B.1: A request need not return useful evidence}
In the jersey-finger case above, six clinically worded requests produce
only four unique facts. The third asks again about appearance and range
of motion but returns the already observed distal-joint tenderness
(\texttt{E6}); the sixth asks about fever and injury history but has no
match. The fourth asks for blood count and inflammatory markers yet the
deterministic matcher returns blood pressure (\texttt{E2}). This is
counted as a new canonical fact under the frozen rule, while its content
does not answer the laboratory question. The trace therefore demonstrates
both why nominal request count overstates evidence yield and why a
\emph{new} status alone is not an information-value metric.
\end{clinicalcase}

\subsection{Result-blind menu controls free-text targeting}
\label{app:b2}

The structured control uses a fixed global library of 103 clinical target
labels. On each turn, the model sees the initial presentation, observed
acquired facts, and target \emph{names} in a case-and-round-seeded
permutation; it does not see target results or the case's complete fact
inventory. Its exact system instruction begins: \emph{``You are selecting
the single next clinical evidence target prospectively. Choose exactly one
label from the global result-blind menu. You see target names but never
hidden results.''} It returns JSON fields \texttt{selected\_target} and
\texttt{reason\_short}. A chosen target's frozen query template and
category are passed through the same deterministic matcher; targets can
be selected again, and a duplicate or no-match still consumes the turn.
The menu experiment is therefore a target-generation interface comparison, not
equalization of evidence sets, realized yields, or prospective information
with Hindsight.

The frozen 120-case primary-menu cohort is a subset of the 181 primary
cases (63 AgentClinic MedQA, 35 dermatology, 22 DiagnosisArena). It uses
the same six budgets and DeepSeek V3.2 diagnostic model. At $k=3$,
free-form and menu accuracy are both 38/120 (31.7\%), while Hindsight is
90/120 (75.0\%). The mean numbers of unique realized facts are 2.09,
1.61, and 3.00, respectively. Across all 720 slots per prospective
condition, free-form yields 412 new, 190 duplicate, and 118 no-match
outcomes; menu yields 263 new, 335 duplicate, and 122 no-match outcomes.
The menu reduces no-match little and increases repetition substantially;
it does not close the diagnostic gap. Accuracy at $k=6$ is 38.3\%
free-form, 42.5\% menu, and 75.8\% Hindsight, with mean unique fact
counts 3.43, 2.19, and 6.00.

\begin{table}[H]
\caption{All six budgets in the 120-case result-blind menu comparison.
Accuracies are top-diagnosis correctness; fact columns are the mean
\emph{unique realized} non-initial facts.}
\label{tab:app-b-menu}
\centering\small
\input{tables/appendix/B/result-blind-menu.tex}
\end{table}

The separate 140-case 12-turn menu continuation uses the same frozen
subset as the A.2 continuation, while reusing an earlier menu prefix only
when the prior prefix was verified as identical for that case. The
DeepSeek V3.2 menu selector and diagnostic probe ran through OpenRouter at
temperature zero, with 1,200- and 2,500-token output limits for new
selection and probe calls; GPT-5.5 judged semantic equivalence. The
following later budgets show that a longer menu run does not eliminate
its repetition bottleneck. The 140-case menu has 53/140 correct top
diagnoses at $k=12$ (37.9\%), compared with 67/140 (47.9\%) free-form
and 108/140 (77.1\%) retrospective. By then it has obtained only 378
new fact events across 1,680 requests, versus 834 free-form; 1,006 menu
requests are duplicates. This continuation is descriptive and shares
cases with the primary analysis.

\begin{table}[H]
\caption{Extended menu comparison on the frozen 140-case subset.
The right columns count realized new and duplicate outcomes cumulatively
over $140k$ request slots per prospective condition; each condition's
diagnostic accuracy is evaluated at the indicated budget.}
\label{tab:app-b-menu12}
\centering\small
\begin{tabular}{rrrrrrrr}
\toprule
 & \multicolumn{3}{c}{Accuracy (\%)} & \multicolumn{2}{c}{Free-form} & \multicolumn{2}{c}{Menu} \\
\cmidrule(lr){2-4}\cmidrule(lr){5-6}\cmidrule(lr){7-8}
$k$ & Free & Menu & Hind & New & Duplicate & New & Duplicate \\
\midrule
3  & 29.3 & 32.1 & 70.0 & 296 & 55  & 230 & 130 \\
6  & 37.1 & 38.6 & 73.6 & 518 & 191 & 306 & 396 \\
9  & 45.7 & 37.9 & 77.9 & 689 & 391 & 351 & 696 \\
12 & 47.9 & 37.9 & 77.1 & 834 & 626 & 378 & 1006 \\
\bottomrule
\end{tabular}
\end{table}

\begin{clinicalcase}{B.2: Repeated menu targets in a lupus case}
For dermatology case \texttt{case\_117}, the initial vignette describes
a 40-year-old woman with a recurrent sun-associated nasal and cheek
eruption; the reference diagnosis is systemic lupus erythematosus. The
menu selector chooses \texttt{EXAM\_SKIN} on each of turns 1--3, and each
template query produces no match. At turn 4 it selects
\texttt{HX\_SKIN}; that history template matches the canonical physical
exam fact \texttt{E1}, eroded butterfly-shaped malar plaques. At turn 5
it requests autoimmune serology, but no such laboratory result is in the
case inventory, so the slot is a no-match. The menu diagnosis is correct
at $k=5$ with one unique returned fact; the free-form trajectory has two
unique facts, including \texttt{E1}, yet remains incorrect in the
diagnostic call. This example exposes both target-to-fact matching limits
and diagnosis-call variability. It cannot establish a patient-level causal
benefit of menu selection, nor does a missing test result mean the test
would be unavailable in real practice.
\end{clinicalcase}

\subsection{Adaptation after an unproductive action}
\label{app:b3}

An adaptation opportunity is a duplicate or no-match at turn
$t\in\{1,\ldots,5\}$ with an observed next request. The unit of analysis
is this \emph{event}, rather than the case; a case can contribute several
events. Recovery is the indicator that the immediately following request
returns a previously unseen canonical fact. The denominator comprises
343 events from 162 cases: 194 duplicates and 149 no-matches. Recovery
occurs after 83/194 duplicates (42.8\%; 95\% CI 35.5--50.6) and
51/149 no-matches (34.2\%; 95\% CI 25.9--44.4). Intervals use 10,000
case-cluster bootstrap draws, seed 20260924, preserving within-case
dependence among events.

\begin{table}[H]
\caption{Immediate recovery by position of the failed request. Entries
are recovered/eligible events (percentage). No duplicate is possible at
turn 1 because no fact has yet been acquired.}
\label{tab:app-b-failure-position}
\centering\small
\begin{tabular}{rcc}
\toprule
Failed turn & After duplicate & After no-match \\
\midrule
1 & --- & 6/17 (35.3\%) \\
2 & 12/25 (48.0\%) & 9/32 (28.1\%) \\
3 & 22/50 (44.0\%) & 18/40 (45.0\%) \\
4 & 24/55 (43.6\%) & 11/31 (35.5\%) \\
5 & 25/64 (39.1\%) & 7/29 (24.1\%) \\
\midrule
All & 83/194 (42.8\%) & 51/149 (34.2\%) \\
\bottomrule
\end{tabular}
\end{table}

Declared request-category switching is a coarse measure of changed
action, not proof that the model explored a different clinical hypothesis.
After a duplicate, 153/194 next requests switch category; after a
no-match, 123/149 switch. Switching is associated with recovery after
76/153 duplicates (49.7\%) versus 7/41 when staying (17.1\%). After a
no-match, the direction reverses: 34/123 switches recover (27.6\%)
versus 17/26 stays (65.4\%). The corresponding case-cluster bootstrap
switch-minus-stay differences are +32.6 pp (95\% CI 17.0--46.3) and
$-37.7$ pp (95\% CI $-56.0$ to $-17.4$). These are observational
associations: target difficulty, case content, turn position, and prior
strategy can confound the comparison.

\begin{table}[H]
\caption{Failure-specific category switching and next-turn recovery.
Rates and 95\% case-cluster bootstrap intervals are taken from the
frozen B.3 analysis.}
\label{tab:app-b-switch}
\centering\small
\input{tables/appendix/new/b3-adaptation.tex}
\end{table}

\begin{table}[H]
\caption{Recovery stratified by the \emph{failed} request's declared
category. Small imaging and medication no-match strata are descriptive.}
\label{tab:app-b-category}
\centering\small
\begin{tabular}{lcc}
\toprule
Failed category & After duplicate & After no-match \\
\midrule
History & 29/82 (35.4\%) & 18/40 (45.0\%) \\
Physical exam & 23/45 (51.1\%) & 16/43 (37.2\%) \\
Laboratory & 15/31 (48.4\%) & 13/34 (38.2\%) \\
Imaging & 7/11 (63.6\%) & 1/11 (9.1\%) \\
Medication & 9/25 (36.0\%) & 1/14 (7.1\%) \\
Other & --- & 2/7 (28.6\%) \\
\bottomrule
\end{tabular}
\end{table}

At case level, the longest run of consecutive nonproductive actions has
length 0 for 10 cases, 1 for 69, 2 for 51, 3 for 22, 4 for 15, 5 for
five, and 6 for nine. Thus 102/181 cases (56.4\%) have a run of at
least two, and 51/181 (28.2\%) a run of at least three. This distribution
uses cases as its denominator and is deliberately separate from the
event-conditioned recovery table.

\begin{clinicalcase}{B.3: Recovery after unproductive requests}
In \texttt{case\_117}, a question about medications and autoimmune
history at turn 2 yields no match, another history question at turn 3
also yields no match, and switching to a physical-exam category at turn 4
still yields no match. Repeating the physical-exam category at turn 5,
this time requesting precise morphology and distribution, returns the
malar-plaque fact \texttt{E1}; a turn-6 serology request again yields no
match. This is a three-failure stagnation run followed by recovery when
the \emph{content} of the request changes without a category switch. In
the jersey-finger case, a duplicate physical-exam result at turn 3 is
followed by a category switch to laboratory at turn 4; the matcher
returns a new blood-pressure fact, although the request asked for blood
count and inflammatory markers. Both traces show why recovery of a new
canonical ID must be distinguished from acquisition of useful clinical
evidence.
\end{clinicalcase}

\paragraph{Scope.}
The B analyses identify losses between action, simulator realization, and
subsequent action. They do not isolate which of lexical matching,
prospective target choice, inaccessible facts, or downstream diagnosis
caused a particular wrong answer. The C and D analyses hold fact count,
fact identity, order, or evidence prefix fixed to address those later
stages of the argument.

%% file: tables/appendix/new/broad-ten-models.tex
\begin{tabular}{lrrrr}
\toprule
Model & $N$ & Full & Fore & Hind \\
\midrule
GPT-5.5 & 89 & 96.4 & 75.3 & 88.8 \\
Gemini 3.1 Pro & 86 & 98.8 & 84.9 & 91.9 \\
DeepSeek V3.2 & 73 & 92.6 & 63.0 & 82.2 \\
Qwen3.8-Max & 82 & 90.9 & 73.2 & 86.6 \\
Claude Sonnet 4.6 & 80 & 90.8 & 70.0 & 82.5 \\
GLM 5.2 & 84 & 88.7 & 81.0 & 81.0 \\
Grok 4.6 & 92 & 85.9 & 78.3 & 90.2 \\
Kimi K2.5 & 80 & 88.6 & 70.9 & 81.0 \\
Llama 4 Maverick & 63 & 84.1 & 71.4 & 88.9 \\
Mistral Large 3 & 69 & 91.3 & 69.6 & 91.3 \\
\bottomrule
\end{tabular}

%% file: tables/appendix/A/primary-paired.tex
\begin{tabular}{rrrrrr}
\toprule
$k$ & Foresight & Hindsight & Gap & 95\% CI low & 95\% CI high \\
\midrule
1 & 18.8\% & 55.8\% & 37.0\% & 29.3\% & 45.3\% \\
2 & 29.3\% & 66.3\% & 37.0\% & 28.7\% & 45.3\% \\
3 & 33.1\% & 73.5\% & 40.3\% & 32.6\% & 48.6\% \\
4 & 35.9\% & 75.1\% & 39.2\% & 31.5\% & 47.0\% \\
5 & 34.8\% & 74.0\% & 39.2\% & 30.9\% & 47.5\% \\
6 & 42.0\% & 75.7\% & 33.7\% & 26.0\% & 41.4\% \\
\bottomrule
\end{tabular}

%% file: tables/appendix/B/result-blind-menu.tex
\begin{tabular}{rrrrrrr}
\toprule
$k$ & Free acc. & Menu acc. & Hindsight acc. & Free facts & Menu facts & Hindsight facts \\
\midrule
1 & 16.7\% & 25.0\% & 57.5\% & 0.89 & 0.90 & 1.00 \\
2 & 26.7\% & 30.0\% & 70.0\% & 1.57 & 1.40 & 2.00 \\
3 & 31.7\% & 31.7\% & 75.0\% & 2.09 & 1.61 & 3.00 \\
4 & 33.3\% & 38.3\% & 76.7\% & 2.58 & 1.81 & 4.00 \\
5 & 31.7\% & 35.8\% & 78.3\% & 3.05 & 2.02 & 5.00 \\
6 & 38.3\% & 42.5\% & 75.8\% & 3.43 & 2.19 & 6.00 \\
\bottomrule
\end{tabular}

%% file: tables/appendix/new/b3-adaptation.tex
\begin{tabular}{llrrr}
\toprule
Failure & Next category & Recovered & Eligible & Rate [95\% CI] \\
\midrule
duplicate & stay & 7 & 41 & 17.1 [6.1, 30.8] \\
duplicate & switch & 76 & 153 & 49.7 [41.2, 58.7] \\
no match & stay & 17 & 26 & 65.4 [47.6, 82.1] \\
no match & switch & 34 & 123 & 27.6 [19.2, 38.3] \\
\bottomrule
\end{tabular}

%% file: sections/appendix/02-C.tex
\section{Experiment C: Acquired Evidence, Identity, Order, and Value}
\label{app:acquired-evidence}

This appendix expands Experiment C in the order of the main text.  The unit of
analysis is the \emph{fact actually returned} by an action, rather than the
request alone.  The diagnosis-unconditioned Hindsight selector sees the
completed case and ranks its non-initial atomic facts, but receives neither the
reference diagnosis nor the prospective trajectory.  It is consequently a
retrospective reference for evidence identity and ordering, not a policy that
could be deployed at the prospective decision point.  The primary paired
cohort contains 181 Fullsight-correct, protocol-complete cases; the six-action
order intervention uses a fixed subset of 100 cases (61 AgentClinic MedQA, 19
dermatology, 20 DiagnosisArena).  The 140-case count-matched continuation is
another fixed subset of the primary cohort.  These denominators are kept
separate throughout.

\paragraph{Prompts, scoring, and deployment.}
The acquisition call asks the clinical policy, using only the initial
presentation, acquired facts, and previous request outcomes, to choose one
free-text clinical information request and a request type.  It cannot see
hidden fact IDs or diagnosis labels.  The environment matcher returns at most
one new atomic fact per action, or records \texttt{duplicate} or
\texttt{no\_match}.  The Hindsight selector is instructed to ``rank exactly six''
non-initial facts by diagnostic usefulness, to use only supplied IDs once, and
is explicitly told that the gold diagnosis, Fullsight diagnosis, and Foresight
trajectory are unavailable.  The same-set reranker instead receives \emph{all
and only} the facts Foresight actually acquired and must use every supplied ID
exactly once.  Fresh diagnostic probes receive just the initial presentation
and the evidence prefix; their JSON output contains a top diagnosis and up to
five ordered differential diagnoses.  The C1 acquisition,
ranking, and probe calls used \texttt{deepseek/deepseek-v3.2} through OpenRouter
at temperature zero, with a six-action cap.  Case/order randomization is
seeded in the frozen run configuration; the provider did not expose a
deterministic generation seed.  Diagnostic correctness uses normalized exact
matching followed where needed by a blinded independent GPT-5.5 judge;
\texttt{uncertain} is scored incorrect.  C1 intervals use paired case-level
percentile bootstraps (5,000 draws), rather than independent provider-call
resampling.

\subsection{Matching the amount actually acquired}
\label{app:c-count}

Let $F_i(k)$ be the set of unique non-initial facts returned to Foresight for
case $i$ after $k$ requests, with $m_i(k)=|F_i(k)|$.  If $H_i$ is the ordered
Hindsight fact list, the count-matched comparator exposes the first $m_i(k)$ facts of
$H_i$ to a fresh diagnostic probe.  Thus the two probes receive the same
\emph{number} of realized facts, while their fact identities and the
selector's information remain different.  The paired accuracy difference is
$N^{-1}\sum_i\{\mathbf 1[\hat y_i(H_i^{1:m_i(k)})=y_i]-
\mathbf 1[\hat y_i(F_i(k))=y_i]\}$.  Empty matched sets stay in the
denominator.  An action producing a duplicate or no match does not increase
$m_i(k)$; no fictitious fact is inserted to fill the slot.

\begin{table}[t]
\centering\small
\caption{Actual fact-count matching.  Accuracy and paired difference are
percentages and percentage points, respectively.  The continuation cohort is
reported separately from the primary 181 cases.}
\label{tab:app-c-count}
\begin{tabular}{lrrrrr}
\toprule
Cohort / requests & Mean $m_i(k)$ & Foresight & Matched Hindsight & Gap & 95\% CI (pp)\\
\midrule
Primary / 3 & 2.09 & 33.1 & 63.0 & 29.8 & [22.1, 37.6]\\
Primary / 6 & 3.52 & 42.0 & 70.7 & 28.7 & [20.4, 37.0]\\
Continuation / 6 & 3.70 & 37.1 & 72.9 & 35.7 & [27.1, 44.3]\\
Continuation / 12 & 5.96 & 47.9 & 69.3 & 21.4 & [12.1, 30.7]\\
\bottomrule
\end{tabular}
\end{table}

The primary count distribution is $11,26,79,65$ cases with $m_i(3)=0,1,2,3$,
and $9,5,23,49,46,39,10$ cases with $m_i(6)=0,1,2,3,4,5,6$.  Table
\ref{tab:app-c-strata} exposes the corresponding casewise comparisons rather
than only their pooled mean.  Small strata have wide intervals and are
descriptive; the matched result is the paired estimate in
Table~\ref{tab:app-c-count}.

\begin{table}[t]
\centering\footnotesize
\caption{Primary-cohort count-matched strata.  $m$ is the number of unique
facts actually obtained.  The last column is the paired Hindsight-minus-Foresight
accuracy difference with a case bootstrap 95\% interval, in percentage points.}
\label{tab:app-c-strata}
\begin{tabular}{rrrrrr}
\toprule
$k$ & $m$ & $N$ & Foresight (\%) & Matched Hindsight (\%) & Gap [95\% CI] (pp)\\
\midrule
3&0&11&0.0&0.0&0.0 [0.0, 0.0]\\
3&1&26&19.2&42.3&23.1 [3.8, 42.3]\\
3&2&79&36.7&74.7&38.0 [25.3, 50.6]\\
3&3&65&40.0&67.7&27.7 [15.4, 40.0]\\
\midrule
6&0&9&0.0&0.0&0.0 [0.0, 0.0]\\
6&1&5&20.0&40.0&20.0 [0.0, 60.0]\\
6&2&23&30.4&56.5&26.1 [4.3, 47.8]\\
6&3&49&46.9&85.7&38.8 [22.4, 55.1]\\
6&4&46&37.0&73.9&37.0 [21.7, 52.2]\\
6&5&39&61.5&76.9&15.4 [-2.6, 33.3]\\
6&6&10&40.0&70.0&30.0 [0.0, 60.0]\\
\bottomrule
\end{tabular}
\end{table}

\begin{clinicalcase}{C.1: A missed functional finding}
In \texttt{agentclinic\_medqa\_14}, a 22-year-old man presents with two days
of right ring-finger pain and swelling; the reference diagnosis is flexor
digitorum profundus tendon avulsion (jersey finger).  By request 3, Foresight
has obtained two unique facts (progressive swelling/pain, \texttt{P5}, and
distal interphalangeal tenderness, \texttt{E6}); its third request repeats
\texttt{E6}.  Hindsight puts the finger caught in another player's jersey
(\texttt{P4}) and inability to flex the distal interphalangeal joint
(\texttt{E7}) first.  Both facts are absent from the prospective six-request
set.  With two Hindsight facts, the fresh diagnostic probe identifies jersey finger;
with two Foresight facts it gives septic arthritis.  The paired correctness
is 1 versus 0 at both $k=3$
and $k=6$.  This case illustrates an identity difference within the same fact
count, rather than proving that either single fact would always be sufficient.
\end{clinicalcase}

\subsection{Critical-fact reach and stable sufficiency}
\label{app:c-identity}

For each case, Critical Evidence Recall at request $k$ is the canonical-ID
overlap between facts in $F_i(k)$ and the first $k$ Hindsight-ranked facts, divided by
the number of Hindsight facts available in that prefix (normally $k$ for $k\leq6$).
Top-1 reach instead asks whether the single highest-ranked Hindsight fact occurs
\emph{anywhere} among facts obtained by request $k$.  They answer distinct
questions: average coverage of a ranked prefix and reach of its first member.
Neither equates the Hindsight ordering with a clinician gold standard.  For the
primary 181 cases, CER@3 is 16.0\%, and Hindsight Top-1 is present in only 18.2\%
by request 3; in 64.6\% it is still absent after six requests.

Stable sufficiency also needs an explicit horizon.  For condition $c$ and
horizon $K=6$, define
\begin{equation}
s_i^{(K)}(c)=\min\{k\in\{0,\ldots,K\}:\hat y_{i,c,j}=y_i
\ \text{for every }j=k,\ldots,K\},
\end{equation}
with $s_i^{(K)}>K$ if the condition never remains correct through $K$.
The reported stable-by-$k$ rate is
$N^{-1}\sum_i\mathbf1[s_i^{(K)}(c)\leq k]$.  It is therefore possible to
be correct at a given prefix but not yet stable.  At $k=2$, the six-step
horizon yields 23.2\% Foresight versus 50.3\% Hindsight.  The primary 181-case
diagnostic trajectory is complete through six requests only; no twelve-step
stable-sufficiency rate is inferred for that primary denominator.  The
fixed 140-case count-matched continuation subset in
Table~\ref{tab:app-c-count} is not used for these stable-sufficiency rates.
Six-request and twelve-request stable-sufficiency horizons are not
interchangeable.

\begin{table}[t]
\centering\footnotesize
\caption{Critical evidence and stable correctness in the primary 181 cases,
requests 1--6.  Stable$^{(6)}$ requires correctness at every later prefix
through request 6.  CER and Top-1 reach use exact canonical fact IDs.  All
entries are percentages.}
\label{tab:app-c-stable6}
\begin{tabular}{rrrrr}
\toprule
$k$ & CER & Top-1 reached & F stable$^{(6)}$ & Hindsight stable$^{(6)}$\\
\midrule
1&5.5&5.5&14.4&43.1\\
2&12.2&13.3&23.2&50.3\\
3&16.0&18.2&27.1&56.4\\
4&20.9&23.8&30.9&60.2\\
5&23.8&30.4&34.8&66.9\\
6&26.9&35.4&42.0&75.7\\
\bottomrule
\end{tabular}
\end{table}

The case above connects reach to sufficiency: \texttt{agentclinic\_medqa\_14}
never reaches Hindsight Top-1 within six requests and is wrong at every prospective
prefix through request 6, while the Hindsight prefix becomes correct at request 2 and
remains correct.  It does not imply that Top-1 reach is necessary in every
case; the cohort rates deliberately report reach and diagnostic stability as
separate outcomes.

\subsection{Changing order while holding the set fixed}
\label{app:c-order}

The direct order experiment uses fresh diagnostic probes for four sequences:
the realized Foresight set in its original unique-fact order (FF), that same
set reranked by the result-blind-to-diagnosis same-set prompt (FR), the Hindsight set
in its Hindsight order (HH), and that same Hindsight set in reverse order (HR).  A sequence
with fewer than six unique facts retains its last available prefix state for
the remaining positions; no new fact is imputed.  The area under the
six-prefix accuracy curve is
$\mathrm{AUSC}_i(c)=\frac16\sum_{k=1}^{6}
\mathbf1[\hat y_{i,c,k}=y_i]$.
The paired contrasts are
$G_{\mathrm{order,F}}=\overline{\mathrm{AUSC}(FR)-\mathrm{AUSC}(FF)}$,
$G_{\mathrm{set}}=\overline{\mathrm{AUSC}(HH)-\mathrm{AUSC}(FR)}$,
and
$S_{\mathrm{order,H}}=\overline{\mathrm{AUSC}(HH)-\mathrm{AUSC}(HR)}$.
The set contrast changes fact identity as well as the retrospective
information available to its selector; it is not a prospectively attainable
causal policy effect.

\begin{table}[t]
\centering\small
\caption{Fixed 100-case set/order intervention.  AUSC is the mean fraction of
the first six prefix diagnoses that are correct; intervals are paired
case-level bootstrap 95\% intervals.  The last three rows report paired
percentage-point contrasts, rather than independent arm comparisons.}
\label{tab:app-c-order}
\begin{tabular}{lrr}
\toprule
Sequence or contrast & Estimate (\%) & 95\% CI\\
\midrule
FF: Foresight set, original order & 34.3 & [26.8, 42.0]\\
FR: Foresight set, reranked & 43.7 & [34.8, 52.7]\\
HR: Hindsight set, reversed & 44.3 & [38.0, 50.8]\\
HH: Hindsight set, Hindsight order & 70.3 & [63.3, 77.0]\\
\midrule
FR $-$ FF: same-set order gain (pp) & 9.3 & [5.5, 13.2]\\
HH $-$ FR: set contrast (pp) & 26.7 & [19.0, 34.5]\\
HH $-$ HR: Hindsight-order sensitivity (pp) & 26.0 & [20.2, 32.0]\\
\bottomrule
\end{tabular}
\end{table}

The case mix makes the direction of these aggregate effects inspectable.
For AgentClinic ($N=61$), dermatology ($N=19$), and DiagnosisArena ($N=20$),
the same-set reranking gains are respectively 9.6, 14.0, and 4.2 points;
the set contrasts are 32.8, 20.2, and 14.2 points; and Hindsight-order sensitivities
are 32.5, 22.8, and 9.2 points.  These subgroup means are exploratory and
their smaller denominators do not support a ranking of source families.

\begin{clinicalcase}{C.3: A skin lesion whose identifying sign was not requested}
In dermatology case \texttt{case\_34}, a 44-year-old woman has a longstanding
left shin nodule; the reference diagnosis is dermatofibroma.  The completed
case contains inward dimpling on pressure (\texttt{E6}), which the Hindsight selector
ranks first.  Foresight instead obtains irritation while shaving
(\texttt{P5}), absence of pain/tenderness (\texttt{P6}), and location on the
shin (\texttt{E5}), with three duplicate-return actions.  The original
prospective order produces no correct diagnosis at any of six prefixes
(FF AUSC 0); reranking only those same three facts makes the probe correct at
prefixes 3--6 (FR AUSC $4/6$).  Hindsight's selected set in Hindsight order gives the
correct diagnosis at all six prefixes (HH AUSC 1), while reversing it delays
correctness until prefix 4 (HR AUSC $3/6$).  The case shows why order can
matter even when the set is unchanged and why ordering the acquired set cannot
recover a high-ranked fact that was never obtained.
\end{clinicalcase}

\begin{figure}[t]
\centering
\includegraphics[width=\linewidth]{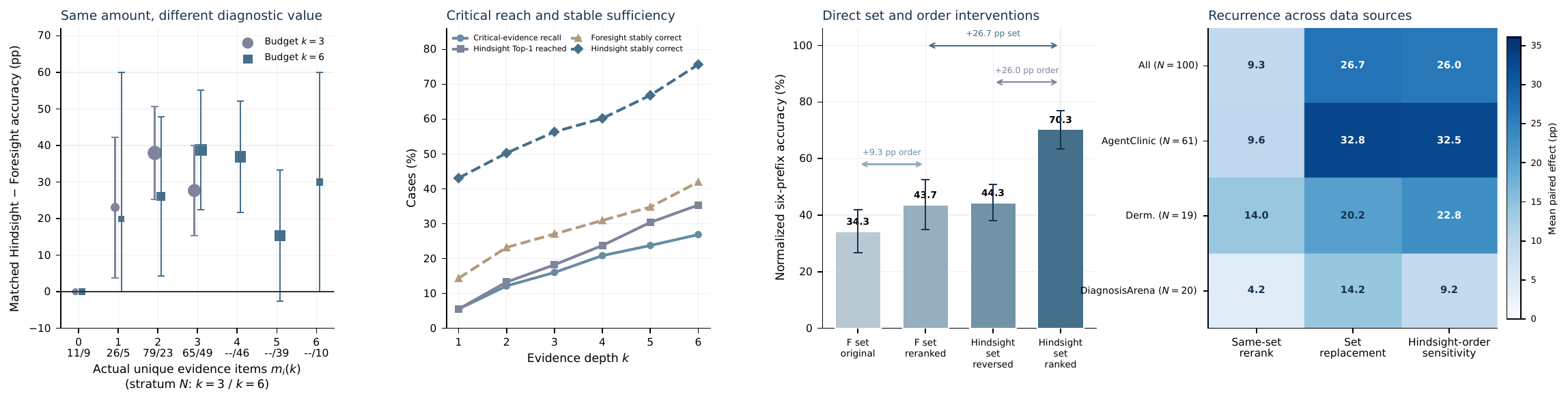}
\caption{Four aligned supplementary panels for Experiment C: residual
diagnostic gap within actual fact-count strata; critical-fact reach and
six-step stable sufficiency; four set/order arms; and source-stratified
effects.  The horizontal four-panel arrangement keeps equally sized panels
at the same visual scale.}
\label{fig:app-c-decomposition}
\end{figure}

\subsection{Distributional change and truth-directed value}
\label{app:c-distribution}

The C2 scatter compares two quantities computed from a fixed candidate
support \emph{within each case}.  For elicited distributions $p$ and $q$ at
two depths, their Jensen--Shannon divergence is
$\mathrm{JSD}(p,q)=\tfrac12\mathrm{KL}_2(p\|m)+
\tfrac12\mathrm{KL}_2(q\|m)$, where $m=(p+q)/2$; it measures update
magnitude in bits, with no direction toward the reference answer.  The
truth-directed coordinate is
$\Delta p_{\mathrm{ref}}=q(y_i)-p(y_i)$; positive values move probability
mass toward the reference diagnosis.  Entropy is
$H(p)=-\sum_jp_j\log_2p_j$; for the separate heatmap experiment it is
normalized by $\log_2 J_i$, where $J_i$ is that case's actual fixed support
size.  Lower entropy alone does not establish better diagnosis.  The
probability vector is an elicited output to an explicit allocation prompt,
not an observation of the model's internal belief.

The six-step scatter comparisons in Table~\ref{tab:app-c-jsd} come from the
\emph{181-case fixed-support current-protocol rereadout}, at endpoints
$0,3,6$.  An interval is included only if at least one new unique fact
appears between its endpoints, so the denominators vary.  The bridge to
older probability calls failed its prespecified agreement gate; the
published six-step comparisons use newly measured states under one current
protocol rather than mixing batches.  In this readout run, the probability allocator
received the current clinical state and a frozen, case-specific set of
anonymized candidate IDs including \texttt{OTHER}; its JSON response had to
assign a probability to every supplied ID and sum to one.  The run used
\texttt{deepseek/deepseek-v3.2} through OpenRouter at temperature zero,
\texttt{reasoning\_effort=low}, JSON-object response format, a 7,500-token
completion cap, and batches of at most eight tasks with no case repeated
inside a batch.  The post-gate amendment renormalized
\emph{all} collected state vectors by their raw sum, with raw vectors
retained; interval means use 10,000 case-level bootstrap draws.  The adjacent
$p_{\mathrm{ref}}$/normalized-entropy heatmaps shown in the main figure come
from a \emph{separate aligned 50-case experiment} with four independent
readouts per state (base twice, candidate order reversed, and rotated).
Those heatmaps and the scatter do not share one 50-case denominator or one
readout protocol.  In the 50-case study, the actual initial-only diagnoses
form a case-specific candidate list, the reference diagnosis is inserted as
an unmarked candidate when needed, and \texttt{OTHER} is appended; candidate
coordinates are aggregated by a median and renormalized.  The formal
collection used DeepSeek V3.2 via OpenRouter, temperature zero,
\texttt{reasoning\_effort=none}, and a 512-token completion cap, with a
10,000-draw nested case/replicate bootstrap.

\begin{table}[t]
\centering\footnotesize
\caption{Six-step C2 scatter inputs from the 181-case rereadout, restricted to
intervals with at least one new fact.  Means and percentile 95\% case-bootstrap
intervals are shown for JSD (bits) and reference-probability change.
Denominators are case--intervals, not a fixed 50-case heatmap sample.}
\label{tab:app-c-jsd}
\begin{tabular}{llrrr}
\toprule
Sequence & Interval & $n$ & Mean JSD [95\% CI] & Mean $\Delta p_{\rm ref}$ [95\% CI]\\
\midrule
Foresight & $0\to3$ & 170 & 0.531 [0.472, 0.586] & 0.084 [0.043, 0.127]\\
Foresight & $3\to6$ & 146 & 0.509 [0.449, 0.572] & 0.099 [0.040, 0.157]\\
Hindsight & $0\to3$ & 181 & 0.633 [0.578, 0.686] & 0.298 [0.239, 0.359]\\
Hindsight & $3\to6$ & 181 & 0.477 [0.420, 0.536] & 0.053 [-0.015, 0.122]\\
\bottomrule
\end{tabular}
\end{table}

The main contrast is at $0\to3$: mean JSD differs by about 0.103 bits,
whereas mean truth-directed change differs by about 0.214 in probability.
That is a descriptive contrast in the elicited distributions.  It does not
identify the expected information gain of an action or a latent belief
transition.  Same-evidence negative controls show
that elicited probability vectors can change even without a new fact.
Individual scatter points therefore need particular caution; aggregate
contrasts and the new-fact restriction are more defensible.  The separate 50-case experiment
also documented candidate-order and repeated-readout variation.  No
per-case probability from one protocol is combined numerically with the
other protocol.

\begin{clinicalcase}{C.4: Change without movement toward the diagnosis}
For \texttt{agentclinic\_medqa\_108}, the initial complaint is a year of
dyspnea in a 60-year-old man, and the reference diagnosis is silicosis.  By
prospective request 3, the acquired facts are absence of recent fever,
chest pain, wheeze, or cough (\texttt{P7}) and worsening with exertion
(\texttt{P4}); the occupational and radiographic findings have not arrived.
The 181-case scatter records a large $0\to3$ Foresight JSD of 0.855
bits, yet $\Delta p_{\mathrm{ref}}=-0.050$.  Hindsight's first three ranked facts
are eggshell lymph-node calcification (\texttt{E2}), interstitial
reticulonodular infiltrates (\texttt{E1}), and prolonged silica exposure
(\texttt{P8}); its corresponding JSD is 0.759 bits and
$\Delta p_{\mathrm{ref}}=+0.850$.  The signs and probabilities are from the
same frozen 181-case scatter protocol.  This example makes the distinction
between update magnitude and direction tangible; it is not an individual
causal estimate.
\end{clinicalcase}

\paragraph{Claim boundary.}
Count matching fixes quantity but not identity; same-set reranking fixes
identity but changes order; set replacement uses retrospective information;
and probability allocation adds a measurement layer with observable
instability.  Together the accuracy-based controls support an acquired-fact
composition and timing deficit.  They do not establish clinical safety,
an optimal prospective strategy, or direct access to an internal diagnostic
state.

%% file: sections/appendix/03-D.tex
\section{Experiment D: Diagnostic updating after evidence arrives}
\label{app:after-evidence}
The D analyses use three distinct estimands: observed transitions after a naturally acquired fact, elicited distributions at frozen evidence prefixes, and immediate diagnostic rescue after an inserted fact. Hindsight is a retrospective reference, not an online policy.

\subsection{Observed state transitions}
\label{app:d-transitions}
The $181$ primary cases have five adjacent transitions across six diagnostic prefixes. Each transition is wrong-to-wrong (WW), wrong-to-correct (WC), correct-to-correct (CC), or correct-to-wrong (CW), using independent diagnosis-equivalence grades. Table~\ref{tab:app-d-transitions} reports all event denominators. Of 630 transitions beginning wrong, 572 remain wrong (90.8\%); 259/275 transitions beginning correct stay correct (94.2\%). Wrong-state persistence is 272/330 when a new fact arrives and 300/300 when the next action is stagnant. New evidence is therefore necessary for a correct transition in these observed rows, but is not sufficient. In 95/181 cases (52.5\%), all diagnoses at $k=1,\ldots,6$ are wrong. The case-mean action-stagnation rate is 41.3\%; next-action recovery after a duplicate or no-match is 134/343 (39.1\%).

\begin{table}[htbp]\centering\small
\caption{Primary six-step trajectory dynamics. ECNR conditions on an initially wrong state followed by a newly arrived hindsight Top-$r$ fact. Intervals use 10,000 case-cluster bootstrap draws, seed 20260920.}
\label{tab:app-d-transitions}
\input{tables/appendix/D/trajectory-dynamics.tex}
\end{table}
These event rates are clustered within cases and are not independent-binomial estimates.

\subsection{Four-interval fixed-support readout}
\label{app:d-readout}
A current-protocol batch remeasured the same 181 cases at a common $k=0$ and at $k=3,6,9,12$ under each evidence path: $181(1+2\times4)=1{,}629$ states. The within-case candidate support includes OTHER and remains fixed. All 1,629 states are structurally valid. A 30-case bridge failed the prespecified agreement gates against historical $k=0,3,6$ data (Top-1 0.440, reference-probability Spearman 0.540, entropy Spearman 0.082, median JSD 0.388 bits); the four-interval estimates therefore use only the current batch.

\begin{table}[htbp]\centering\footnotesize
\caption{Mean change [case-bootstrap 95\% CI] in elicited reference-diagnosis probability, $N=181$ per interval. The last column is the Hindsight-minus-Foresight difference in changes.}
\label{tab:app-d-prob}
\input{tables/appendix/new/d1-probability.tex}
\end{table}
\begin{table}[htbp]\centering\footnotesize
\caption{Mean change [case-bootstrap 95\% CI] in elicited entropy, in bits, from the same current-protocol states. A decrease means greater concentration, regardless of diagnostic correctness.}
\label{tab:app-d-ent}
\input{tables/appendix/new/d1-entropy.tex}
\end{table}
\begin{figure}[htbp]\centering
\includegraphics[width=\linewidth]{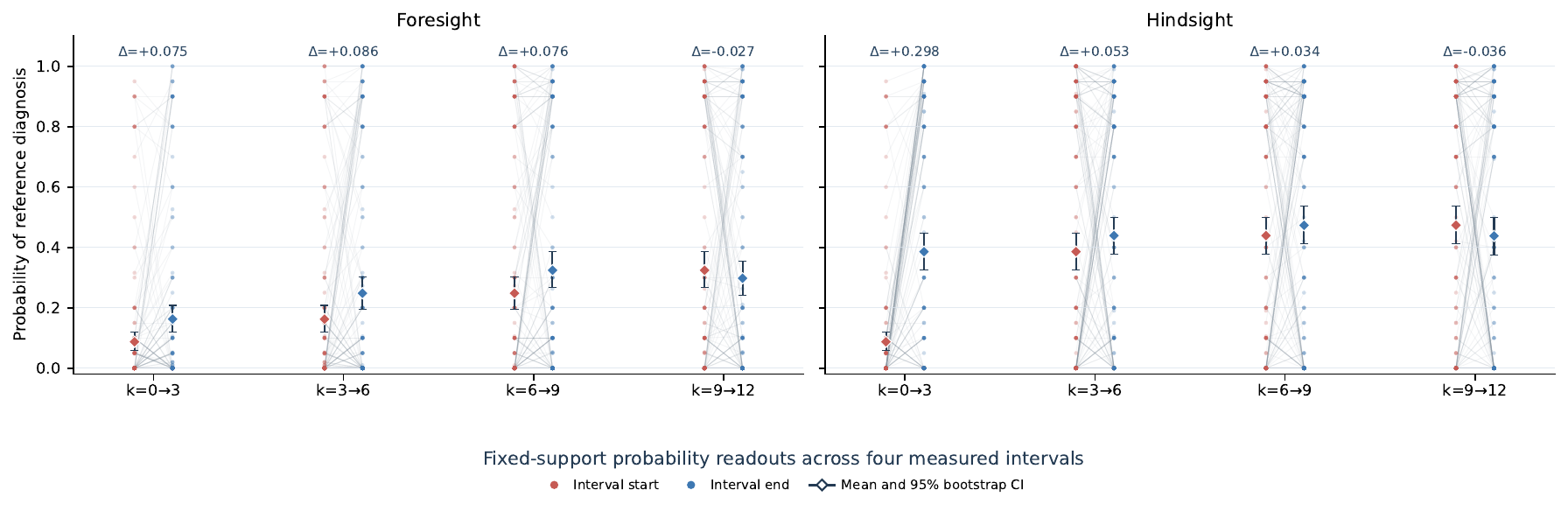}
\par\vspace{1pt}
\includegraphics[width=\linewidth]{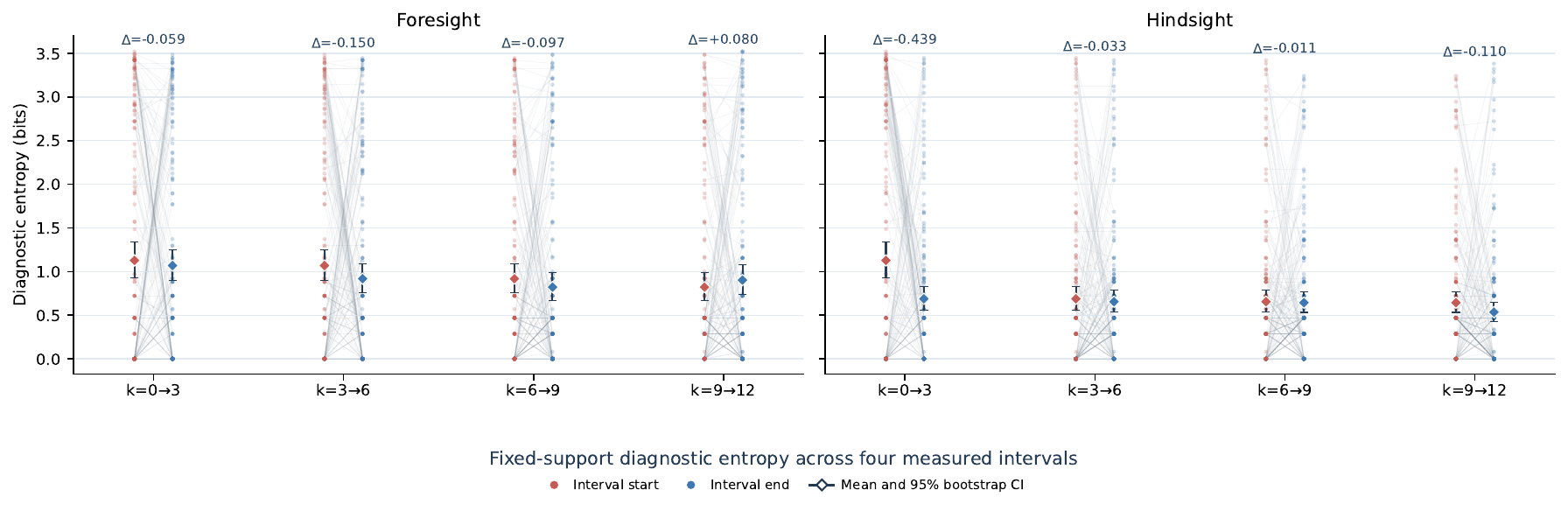}
\caption{Detailed current-protocol state-pair readouts for the 181 cases.
The upper two panels show reference-diagnosis probability and the lower
two show diagnostic entropy, each separated by Foresight and Hindsight.
Faint lines represent individual elicited state pairs; diamonds and
intervals summarize endpoint means. The main figure reports the more
legible interval changes from these same states.}
\label{fig:app-d-statepairs}
\end{figure}
Over the first three request steps, Hindsight raises mean $p_{\rm ref}$ by 0.298 versus 0.075 for Foresight; the difference in changes is 0.223 [0.153, 0.293]. Hindsight entropy falls by 0.439 bits, whereas the Foresight change is -0.059 [-0.320, 0.200]. Later between-condition contrasts have intervals crossing zero. These are elicited probability readouts, not direct observations of latent model belief.

The same deterministic projection $p'_j=p_j/\sum_\ell p_\ell$ was applied to all structurally valid current-batch vectors. Before projection, 28/1,629 vectors in 24 cases had raw mass differing from one by more than 0.01 (range 0.90--1.10). Excluding all 24 affected cases leaves the first hindsight interval at +0.294 [0.230, 0.360]. Identical-evidence negative controls have 37.5--52.9\% Top-1 changes and mean JSD 0.37--0.45 bits. Therefore individual-state lines illustrate heterogeneity, and small changes within one case should not be treated as causal evidence effects.

\subsection{Rank-by-timing rescue}
\label{app:d-rank-rescue}
The main D.2 experiment starts from the fixed 140-case twelve-step subset of the primary cohort. Its fixed complete grid contains 33 cases eligible for all 15 cells: Hindsight rank $r\in\{1,2,3\}$ crossed with insertion after Atomic step $k\in\{1,3,6,9,12\}$. At every cell, the critical and control probes share the same initial presentation and Atomic prefix; only the appended fact changes. Frozen controls are unseen through step 12, outside Hindsight ranks 1--6, and selected by a documented broad-modality/inventory fallback. This is not information-equivalent matching. A repeated unaugmented prefix estimates output variability. The endpoint is correctness immediately after insertion.

\begin{table}[htbp]\centering\scriptsize
\caption{Fixed complete-grid rescue ($N=33$ in every cell). Critical and control columns are percent correct; difference is percentage points [case-cluster 95\% CI].}
\label{tab:app-d-balanced}
\input{tables/appendix/new/d2-balanced-grid.tex}
\end{table}
Across the five depths, mean critical-minus-control rescue is +38.8, +24.8, and +10.3 points for ranks 1, 2, and 3. The Rank-1-minus-Rank-3 contrast is +28.5 [10.9, 46.7], Holm-adjusted $p=0.0163$ after 200,000 sign-flip permutations. The Rank-1 $k=1$ versus $k=12$ contrast is +9.1 [-3.0, 21.2], adjusted $p=0.3747$; the rank-by-time interaction is +18.2 [0.0, 36.4], adjusted $p=0.2202$. The fixed-grid data support a rank effect more clearly than a timing effect.

\begin{table}[htbp]\centering\scriptsize
\caption{All-eligible sensitivity. Every critical-minus-control point estimate is positive, but the denominator changes by rank and depth; these rows are not a fixed-cohort timing series.}
\label{tab:app-d-eligible}
\input{tables/appendix/new/d2-all-eligible.tex}
\end{table}
\begin{figure}[htbp]\centering
\includegraphics[width=\linewidth]{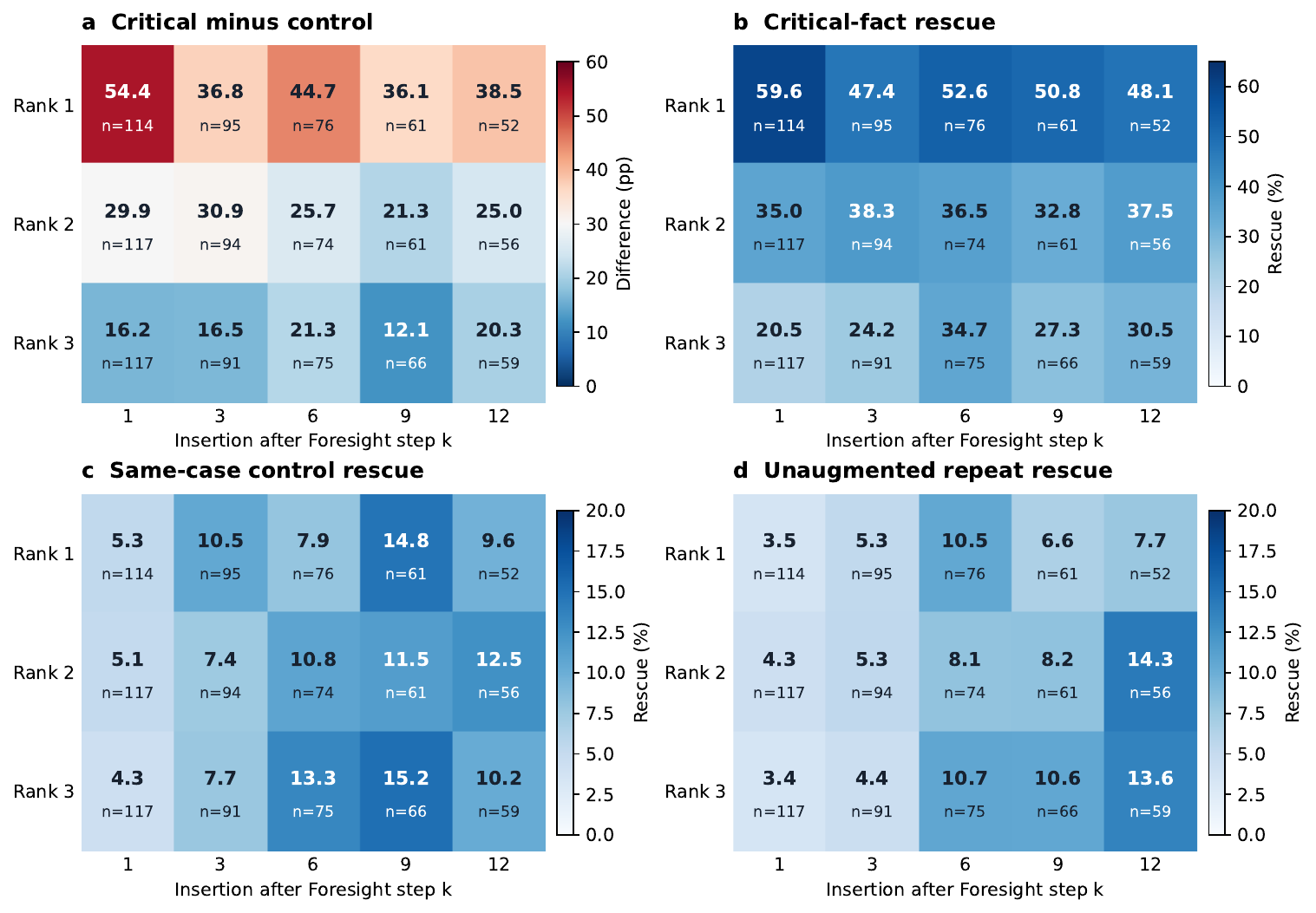}
\caption{Four panels for the all-eligible D.2 sensitivity analysis:
critical-minus-control rescue in percentage points, critical-fact rescue,
pre-frozen same-case control rescue, and unaugmented repeat rescue.
Each cell prints its eligible count; denominators vary across ranks and
insertion depths and cannot be read as a fixed-cohort timing series.
The control follows the coarse modality/inventory rule.}
\label{fig:app-d-eligible}
\end{figure}
Natural-arrival nonresponse is 15/38 for Rank 1, 33/53 for Ranks 2--3, and 58/68 for Ranks 4--6. Frozen correctness retests agree in 89.4\% of states, but normalized exact Top-1 strings agree in 41.6\%; the sign of the critical-minus-control effect agrees in 12/15 cells.

\begin{clinicalcase}{D.2: Chronic cough and rank-dependent rescue}
Case \texttt{agentclinic\_medqa\_151} is a 50-year-old man with two months of cough; the reference diagnosis is Valley fever. Hindsight Rank 1 is pathology describing necrotizing granulomas with spherules and endospores (E5); Rank 2 is hiking near Phoenix (P11); Rank 3 is positive PAS/silver staining (E10). The respective controls are no joint deformity (E20), occasional alcohol use (P10), and blood pressure 120/82 (E13), matched only on recorded coarse modality. The unaugmented repeat predicts upper airway cough syndrome and is incorrect at all five depths. Appending E5 yields coccidioidomycosis and is correct in 5/5 probes; appending P11 is also correct in 5/5; E10 instead leads to an incorrect Pneumocystis diagnosis in 5/5. All fifteen controls remain incorrect. This illustration does not estimate a population effect.
\end{clinicalcase}

\subsection{Distinct single-fact Atomic@3 rescue}
\label{app:d-single-rescue}
A second, separately frozen intervention contains 114 primary cases that were wrong at Atomic@3, had not acquired Hindsight Top-1, and had a complete diagnostic probe. S0 is the original wrong result. S1 appends the missing Hindsight Top-1; S2 appends one frozen unseen same-case control outside Hindsight Top-3, preferring the same recorded modality and using a documented fallback otherwise. Both use the identical Atomic@3 prefix, with the added fact last. These 114 cases are not the 33 complete-grid cases and the effects must not be pooled.
\begin{table}[htbp]\centering\footnotesize
\caption{Frozen $N=114$ single-fact rescue. The difference is paired percentage points; the worst-case row conservatively scores two transport-salvaged outputs.}
\label{tab:app-d-single-rescue}
\input{tables/appendix/D/critical-evidence-rescue.tex}
\end{table}
\begin{table}[htbp]\centering\footnotesize
\caption{Exploratory single-fact strata by clean-prefix sufficiency and control-modality fallback.}
\label{tab:app-d-single-strata}
\input{tables/appendix/D/critical-evidence-rescue-sensitivity.tex}
\end{table}
S1 rescues 19/114 (16.7\%), versus 6/114 (5.3\%) for S2: paired +11.4 points [4.4, 19.3], exact McNemar $p=0.0044$. In the 53 cases where Hindsight Top-1 alone diagnoses correctly from a clean prefix, the S1-S2 contrast is +24.5 points; it is 0.0 in the other 61. This post-hoc split describes heterogeneity and does not identify a unique mechanism.

\begin{clinicalcase}{D.3: Recurrent lesions in a child}
In \texttt{agentclinic\_medqa\_114}, a three-year-old has recurrent skin lesions since infancy. Atomic@3 predicts atopic dermatitis. Adding missed Hindsight Top-1 E4 (abnormal nitroblue tetrazolium test) changes the diagnosis to chronic granulomatous disease, whereas frozen control E9 (respiratory-rate record) leaves the atopic-dermatitis prediction. This is an observed diagnostic intervention, not evidence that a prospective agent knew to request E4.
\end{clinicalcase}

\subsection{Secondary readout and value analyses}
\label{app:d-secondary}
Readout robustness uses 120 isolated states: 40 initial, 40 Atomic@3, and 40 Hindsight@3. Each has identical repeats, candidate-order permutations, and a paraphrased prompt; a separate audit replaces low-probability candidates. Moderate entropy stability and support sensitivity rule out treating any one vector as a validated internal belief. Descriptive calibration across 120 states gives 10-bin ECE 0.135, Brier 0.872, and log loss 3.631.
\begin{table}[htbp]\centering\footnotesize
\caption{Elicited-readout reliability. Top-1 is agreement; probability and entropy columns are Spearman correlations; JSD is median Jensen--Shannon divergence.}
\label{tab:app-d-reliability}
\input{tables/appendix/D/probability-readout.tex}
\end{table}
A bounded local fact panel spans 200 $t=0,3$ decision points from 100 cases, but 138 points in 92 cases have a matched actual choice and at least two valid candidates. Mean actual-choice value percentile is 0.401 [0.350, 0.453]; only 7.2\% choose the highest-TDG candidate. The separate free/menu/random policy comparison below has different estimability subsets and should not be equated with that estimate. TDG is a realized log-probability change, not prospective expected information gain. Changing the reference-probability floor from $10^{-6}$ to $10^{-9}$ changes some mean TDG values by as much as 1.156 bits. This analysis does not support preferential selection of the highest-value candidate.
\begin{table}[htbp]\centering\footnotesize
\caption{Local choice-value comparison. CVP is choice-value percentile; selected TDG is in bits.}
\label{tab:app-d-value}
\input{tables/appendix/D/value-analysis.tex}
\end{table}
\begin{figure}[htbp]\centering
\includegraphics[width=\linewidth]{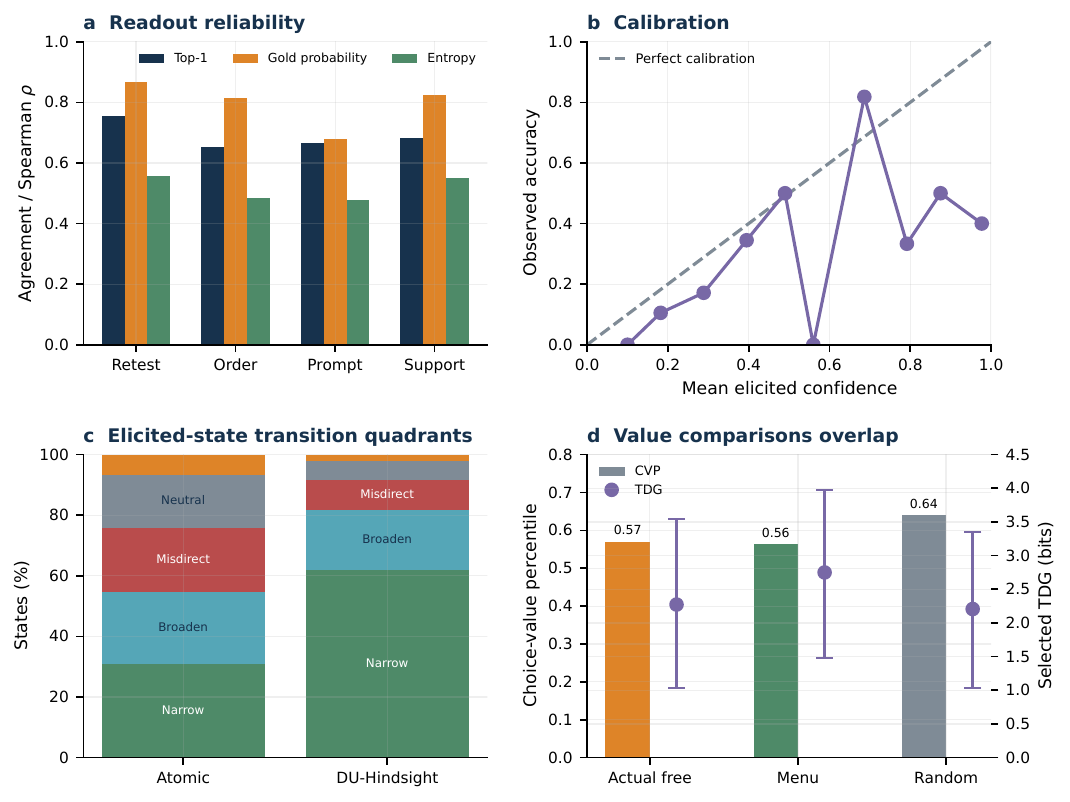}
\caption{Four aligned secondary panels: readout reliability, calibration, elicited-state transition classes, and local value-policy comparison. These analyses are sensitivity evidence.}
\label{fig:app-d-secondary}
\end{figure}

%% file: tables/appendix/D/trajectory-dynamics.tex
% Auto-generated by analyze_trajectory_dynamics.py; do not edit.
\begin{tabular}{lrrl}
\toprule
Measure & Numerator & Denominator & Estimate (95\% CI) \\
\midrule
Wrong-state persistence & 572 & 630 & 90.8\% [88.2, 93.1] \\
Wrong-to-correct transition & 58 & 630 & 9.2\% [6.9, 11.8] \\
Correct-state retention & 259 & 275 & 94.2\% [90.9, 96.9] \\
Action stagnation rate & -- & 181 cases & 41.3\% [37.9, 44.8] \\
Failure recovery by next action & 134 & 343 & 39.1\% [33.2, 45.4] \\
ECNR@1 & 15 & 38 & 39.5\% [23.9, 55.6] \\
ECNR@3 & 48 & 91 & 52.7\% [42.4, 62.7] \\
ECNR@6 & 106 & 159 & 66.7\% [58.9, 74.0] \\
\bottomrule
\end{tabular}

%% file: tables/appendix/new/d1-probability.tex
\begin{tabular}{lrrr}
\toprule
Interval & Foresight & Hindsight & Hind $-$ Fore \\
\midrule
$0\rightarrow3$ & 0.075 [0.036, 0.116] & 0.298 [0.239, 0.359] & 0.223 [0.153, 0.293] \\
$3\rightarrow6$ & 0.086 [0.037, 0.135] & 0.053 [-0.016, 0.121] & -0.032 [-0.120, 0.054] \\
$6\rightarrow9$ & 0.076 [0.024, 0.131] & 0.034 [-0.030, 0.100] & -0.042 [-0.122, 0.038] \\
$9\rightarrow12$ & -0.027 [-0.088, 0.032] & -0.036 [-0.106, 0.032] & -0.009 [-0.101, 0.080] \\
\bottomrule
\end{tabular}

%% file: tables/appendix/new/d1-entropy.tex
\begin{tabular}{lrrr}
\toprule
Interval & Foresight & Hindsight & Hind $-$ Fore \\
\midrule
$0\rightarrow3$ & -0.059 [-0.320, 0.200] & -0.439 [-0.680, -0.199] & -0.380 [-0.609, -0.155] \\
$3\rightarrow6$ & -0.150 [-0.395, 0.093] & -0.033 [-0.217, 0.154] & 0.117 [-0.201, 0.429] \\
$6\rightarrow9$ & -0.097 [-0.310, 0.117] & -0.011 [-0.179, 0.157] & 0.085 [-0.206, 0.365] \\
$9\rightarrow12$ & 0.080 [-0.131, 0.293] & -0.110 [-0.269, 0.051] & -0.190 [-0.462, 0.086] \\
\bottomrule
\end{tabular}

%% file: tables/appendix/new/d2-balanced-grid.tex
\begin{tabular}{rrrrrr}
\toprule
Hindsight rank & $k$ & $N$ & Critical & Control & Difference [95\% CI] \\
\midrule
1 & 1 & 33 & 48.5 & 3.0 & 45.5 [27.3, 63.6] \\
1 & 3 & 33 & 39.4 & 3.0 & 36.4 [18.2, 54.5] \\
1 & 6 & 33 & 45.5 & 6.1 & 39.4 [21.2, 57.6] \\
1 & 9 & 33 & 45.5 & 9.1 & 36.4 [15.2, 57.6] \\
1 & 12 & 33 & 42.4 & 6.1 & 36.4 [18.2, 54.5] \\
2 & 1 & 33 & 27.3 & 9.1 & 18.2 [0.0, 36.4] \\
2 & 3 & 33 & 39.4 & 6.1 & 33.3 [15.2, 51.5] \\
2 & 6 & 33 & 33.3 & 6.1 & 27.3 [9.1, 45.5] \\
2 & 9 & 33 & 24.2 & 6.1 & 18.2 [3.0, 33.3] \\
2 & 12 & 33 & 33.3 & 6.1 & 27.3 [9.1, 45.5] \\
3 & 1 & 33 & 12.1 & 6.1 & 6.1 [0.0, 15.2] \\
3 & 3 & 33 & 15.2 & 6.1 & 9.1 [0.0, 21.2] \\
3 & 6 & 33 & 24.2 & 9.1 & 15.2 [-3.0, 33.3] \\
3 & 9 & 33 & 15.2 & 9.1 & 6.1 [0.0, 15.2] \\
3 & 12 & 33 & 21.2 & 6.1 & 15.2 [3.0, 27.3] \\
\bottomrule
\end{tabular}

%% file: tables/appendix/new/d2-all-eligible.tex
\begin{tabular}{rrrrrr}
\toprule
Hindsight rank & $k$ & $N$ & Critical & Control & Difference [95\% CI] \\
\midrule
1 & 1 & 114 & 59.6 & 5.3 & 54.4 [44.7, 64.0] \\
1 & 3 & 95 & 47.4 & 10.5 & 36.8 [25.3, 48.4] \\
1 & 6 & 76 & 52.6 & 7.9 & 44.7 [32.9, 56.6] \\
1 & 9 & 61 & 50.8 & 14.8 & 36.1 [21.3, 50.8] \\
1 & 12 & 52 & 48.1 & 9.6 & 38.5 [23.1, 53.8] \\
2 & 1 & 117 & 35.0 & 5.1 & 29.9 [20.5, 39.3] \\
2 & 3 & 94 & 38.3 & 7.4 & 30.9 [21.3, 41.5] \\
2 & 6 & 74 & 36.5 & 10.8 & 25.7 [12.2, 37.8] \\
2 & 9 & 61 & 32.8 & 11.5 & 21.3 [8.2, 34.4] \\
2 & 12 & 56 & 37.5 & 12.5 & 25.0 [10.7, 39.3] \\
3 & 1 & 117 & 20.5 & 4.3 & 16.2 [9.4, 23.1] \\
3 & 3 & 91 & 24.2 & 7.7 & 16.5 [8.8, 25.3] \\
3 & 6 & 75 & 34.7 & 13.3 & 21.3 [10.7, 33.3] \\
3 & 9 & 66 & 27.3 & 15.2 & 12.1 [4.5, 19.7] \\
3 & 12 & 59 & 30.5 & 10.2 & 20.3 [8.5, 32.2] \\
\bottomrule
\end{tabular}

%% file: tables/appendix/D/critical-evidence-rescue.tex
% Auto-generated by analyze_critical_fact_rescue.py; do not edit.
\begin{tabular}{lrrl}
\toprule
Condition/contrast & Numerator & N & Estimate (95\% CI) \\
\midrule
S1: add missing Hindsight Top-1 & 19 & 114 & 16.7\% [9.6, 23.7] \\
S2: add random unseen control & 6 & 114 & 5.3\% [1.8, 9.6] \\
Paired S1--S2 & 13 & 114 & 11.4 pp [4.4, 19.3] \\
Worst-case S1 (2 salvages incorrect) & 18 & 114 & 15.8\% [9.6, 22.8] \\
\bottomrule
\end{tabular}

%% file: tables/appendix/D/critical-evidence-rescue-sensitivity.tex
% Auto-generated by analyze_critical_fact_rescue.py; do not edit.
\begin{tabular}{lrrrl}
\toprule
Subset & S1 correct & S2 correct & N & Paired difference (95\% CI) \\
\midrule
Hindsight Top-1 prefix correct & 18 & 5 & 53 & 24.5 pp [11.3, 37.7] \\
Hindsight Top-1 prefix incorrect & 1 & 1 & 61 & 0.0 pp [-4.9, 4.9] \\
Control modality matched & 12 & 4 & 83 & 9.6 pp [2.4, 18.1] \\
Control fallback modality & 7 & 2 & 31 & 16.1 pp [0.0, 32.3] \\
\bottomrule
\end{tabular}

%% file: tables/appendix/D/probability-readout.tex
\begin{tabular}{lrrrr}
\toprule
Perturbation & Top-1 & $\rho$(gold) & $\rho$(entropy) & Median JSD \\
\midrule
Test-retest & 75.4\% & 0.867 & 0.556 & 0.081 \\
Candidate order & 65.4\% & 0.814 & 0.485 & 0.088 \\
Prompt paraphrase & 66.7\% & 0.680 & 0.479 & 0.093 \\
Candidate-support replacement & 68.3\% & 0.823 & 0.550 & 0.100 \\
\bottomrule
\end{tabular}

%% file: tables/appendix/D/value-analysis.tex
\begin{tabular}{lrrr}
\toprule
Choice policy & Estimable & Mean CVP & Mean selected TDG \\
\midrule
Actual free & 138 & 0.570 & 2.274 \\
Prospective menu & 147 & 0.565 & 2.750 \\
Random panel & 200 & 0.639 & 2.207 \\
\bottomrule
\end{tabular}

%% file: sections/appendix/04-E.tex
\section{Experiment E: Robustness, recurrence, and external transfer}
\label{app:robustness}

\subsection{Selector repeatability and downstream sensitivity}
The primary Hindsight selector sees the complete realized non-initial fact pool but neither the reference diagnosis nor Fullsight output. The selector audit freezes 100 cases, makes three independent repeats, two fact-order permutations, and one prompt paraphrase, and separately tests GPT-5.5 and Gemini 3.1 Pro Preview as alternative selectors on 80 cases. In the three-repeat pairwise distribution, mean Top-1 agreement is 83.7\%, Top-3 Jaccard is 0.789, and rank-biased overlap at six is 0.862. The selector is therefore not deterministic at the fact level. Table~\ref{tab:app-e-selector} reports each variant's agreement and downstream Hindsight@3 diagnostic accuracy; positive retrospective utility survives the tested perturbations, while exact ranks vary.

\begin{table}[htbp]\centering\footnotesize
\caption{Selector perturbation audit. Agreement is against the primary selector; Hindsight@3 is the resulting diagnostic accuracy, not a selector accuracy score. Alternative-model rows have $N=80$; same-model perturbations have $N=100$.}
\label{tab:app-e-selector}
\input{tables/appendix/E/selector-robustness.tex}
\end{table}
The candidate pool, IDs, diagnostic probe, and clinical label are held fixed while the ranking call changes. The first two facts in a perturbed ranking may differ from the primary ranking, so this is a sensitivity test of the claim, not evidence of a unique ground-truth evidence order.

\subsection{Shared-case model and source recurrence}
The paired four-model replication uses the identical 59 eligible cases for DeepSeek V3.2, GPT-5.5, Gemini 3.1 Pro Preview, and GLM 5.2. The acquisition/diagnosis roles use the model assigned to each condition; the diagnosis-equivalence judge is separate. The gap is positive in all four point estimates. Gemini's 95\% interval includes zero, so this is directional recurrence, not a claim that every backbone has a statistically resolved gap.

\begin{table}[htbp]\centering\footnotesize
\caption{Paired three-action recognition-acquisition gap on the same 59 cases per model. The gap interval resamples shared cases and is not a model ranking.}
\label{tab:app-e-models}
\input{tables/appendix/E/cross-model.tex}
\end{table}

A separate shared 50-case menu control gives the following free/menu/Hindsight accuracies at three actions: DeepSeek 40/52/60\%, Gemini 56/58/70\%, GLM 40/48/66\%, and GPT-5.5 52/60/68\%. This subset asks a different interface question from the 59-case paired replication and is not pooled with it. Within the 181-case primary cohort, the three source strata at $k=3$ are AgentClinic MedQA ($N=98$): Foresight 34.7\%, Hindsight 80.6\%, gap 45.9 points; dermatology ($N=49$): 42.9\%, 85.7\%, gap 42.9; DiagnosisArena ($N=34$): 14.7\%, 35.3\%, gap 20.6. These are conditional-on-Fullsight-correct source comparisons; the small DiagnosisArena eligible stratum limits precision.

\subsection{EviMed/REFINE bridge}
\label{app:e-evimed}
The external bridge uses 200 cases, 50 each from AgentClinic MedQA, DiagnosisArena, RareArena, and Derm/CRAFT-MD, after source-aware ID, normalized-text-hash, and high-similarity overlap audits against the internal campaign. Doctor, patient/reporter, summarizer, diagnostician, and verifier use DeepSeek V3.2 at temperature zero. The official baseline and REFINE retain their official roles, action space, indexed-fact patient/reporter response rules, stopping rule, and summarizer-diagnostician-verifier loop. A compatibility adapter multiplexes independent cases through a JSON transport envelope; it changes transport, not case-local visibility. The global 103-target result-blind menu is an additional interface control. The Hindsight path remains a retrospective, non-interactive reference.

\begin{table}[htbp]\centering\footnotesize
\caption{EviMed at the official 16-turn budget, $N=200$ per strategy. ICR is indexed-reference information coverage; diagnostic accuracy is independently graded; facts are mean unique realized items. The Hindsight row is retrospective and acquires more facts.}
\label{tab:app-e-evimed}
\input{tables/appendix/E/evimed.tex}
\end{table}
\begin{table}[htbp]\centering\footnotesize
\caption{EviMed prefix endpoints. The rows at 3 and 6 turns are descriptive; the official evaluation budget is 16 turns. ICR and accuracy entries are percentages.}
\label{tab:app-e-prefix}
\input{tables/appendix/new/evimed-prefixes.tex}
\end{table}
At 16 turns, REFINE exceeds the official free-form baseline by 17.9 ICR points (46.2\% versus 28.3\%) and 14.5 accuracy points (38.0\% versus 23.5\%). REFINE also obtains 9.48 versus 5.74 unique facts on average. The comparison therefore establishes that a stronger prospective workflow can improve realized coverage and diagnosis, but it is not an equal-evidence-quantity mechanism test. At the same budget, the menu obtains 6.49 facts and 35.0\% accuracy; the retrospective Hindsight reference obtains 14.12 and 53.0\%.

\subsection{Interpretation of recurrence}
The 140-case twelve-action continuation is a fixed subset of the strict 181-case cohort; the 59-case model replication and 200-case external bridge answer separate questions with different budgets and interfaces. Their directional results support recurrence of an acquisition problem. They do not provide a pooled clinical accuracy estimate, show that Hindsight is prospectively attainable, or remove simulator and readout limitations.

%% file: tables/appendix/E/selector-robustness.tex
\begin{tabular}{llrrrr}
\toprule
Selector & Variant & Top-1 & Jaccard & RBO@6 & Hindsight@3 \\
\midrule
Gemini & cross selector & 72.5\% & 0.713 & 0.809 & 72.5\% \\
GPT & cross selector & 67.5\% & 0.710 & 0.805 & 73.8\% \\
DS & order 1 & 77.0\% & 0.719 & 0.829 & 70.0\% \\
DS & order 2 & 77.0\% & 0.769 & 0.833 & 70.0\% \\
DS & primary & 100.0\% & 1.000 & 1.000 & 76.0\% \\
DS & prompt alt & 78.0\% & 0.747 & 0.829 & 72.0\% \\
DS & repeat 1 & 79.0\% & 0.769 & 0.854 & 70.0\% \\
DS & repeat 2 & 80.0\% & 0.813 & 0.851 & 71.0\% \\
DS & repeat 3 & 80.0\% & 0.798 & 0.866 & 68.0\% \\
\bottomrule
\end{tabular}

%% file: tables/appendix/E/cross-model.tex
\begin{tabular}{lrrrr}
\toprule
Model & Atomic@3 & Hindsight@3 & Gap & 95\% CI \\
\midrule
DeepSeek V3.2 & 35.6\% & 55.9\% & 20.3\% & [6.8\%, 33.9\%] \\
GPT-5.5 & 47.5\% & 66.1\% & 18.6\% & [10.2\%, 28.8\%] \\
Gemini 3.1 Pro Preview & 54.2\% & 62.7\% & 8.5\% & [-3.4\%, 20.3\%] \\
GLM 5.2 & 37.3\% & 61.0\% & 23.7\% & [11.9\%, 37.3\%] \\
\bottomrule
\end{tabular}

%% file: tables/appendix/E/evimed.tex
\begin{tabular}{lrrrr}
\toprule
Strategy & ICR & Accuracy & Unique facts & No-new rate \\
\midrule
Official baseline & 28.3\% & 23.5\% & 5.74 & 0.0\% \\
Menu control & 32.6\% & 35.0\% & 6.49 & 82.7\% \\
Official REFINE & 46.2\% & 38.0\% & 9.48 & 0.0\% \\
Hindsight reference & 73.3\% & 53.0\% & 14.12 & 0.0\% \\
\bottomrule
\end{tabular}

%% file: tables/appendix/new/evimed-prefixes.tex
\begin{tabular*}{\linewidth}{@{\extracolsep{\fill}}rlrrr@{}}
\toprule
$k$ & Strategy & ICR & Accuracy & Facts \\
\midrule
3 & Official baseline & 16.0 & 16.0 & 2.98 \\
3 & Menu & 23.2 & 19.0 & 4.26 \\
3 & REFINE & 16.5 & 14.0 & 2.93 \\
3 & Hindsight reference & 18.9 & 36.0 & 3.00 \\
6 & Official baseline & 25.2 & 22.0 & 5.01 \\
6 & Menu & 27.1 & 25.5 & 5.20 \\
6 & REFINE & 29.6 & 27.0 & 5.88 \\
6 & Hindsight reference & 36.8 & 42.5 & 5.96 \\
16 & Official baseline & 28.3 & 23.5 & 5.74 \\
16 & Menu & 32.6 & 35.0 & 6.49 \\
16 & REFINE & 46.2 & 38.0 & 9.48 \\
16 & Hindsight reference & 73.3 & 53.0 & 14.12 \\
\bottomrule
\end{tabular*}

%% file: sections/appendix/05-reproducibility.tex
\section{Quality controls, execution, and claim boundaries}
\label{app:reproducibility}

\subsection{Numerical and protocol checks}
The analyses retain case identifiers across eligibility, evidence acquisition, diagnostic readouts, and interventions, allowing denominators to be checked at the case, step, and case--interval levels. The primary six-step analyses use 181 eligible cases; the extended action analyses use the fixed 140-case subset; the diagnostic-menu comparison uses 120 cases; and the complete rank-by-timing intervention grid uses 33 cases. The corresponding all-eligible rank-by-timing analysis has varying denominators and is reported separately as a sensitivity analysis. Bootstrap resampling is clustered by case when a case contributes multiple observations. The reported figures and tables were checked against their stated cohorts, estimands, and uncertainty calculations.

The broad ten-model screening analysis is a separate, descriptive experiment. Its selector could condition on each model's own Fullsight diagnosis, whereas the strict primary Hindsight trajectory does not receive the reference diagnosis or a Fullsight answer. The screen and the primary analysis therefore address different questions. Neither the screen nor a Hindsight ranking should be interpreted as an attainable prospective upper bound.

\subsection{Protocol dates, execution, and deviations}
The primary six-step protocol was frozen on 2026-09-17; its case-bootstrap seed is 20260917. The 114-case single-fact rescue was frozen on 2026-09-19, before S1/S2 outcomes, with seed 20260920 and 10,000 case-bootstrap draws. The D2 rank-by-timing grid and same-batch D1 readout are separate later analyses; D2 uses seed 20260924, 10,000 case-cluster bootstrap draws, 200,000 sign-flip permutations for prespecified contrasts, and Holm correction over its three confirmatory contrasts. The D1 current-batch readout uses seed 20260924 and 10,000 case-bootstrap draws. Provider calls were routed through OpenRouter with temperature zero, but a provider seed was unavailable, so exact replay of model outputs is not guaranteed even with frozen prompts. Request and response checksums support run-level consistency checks rather than deterministic replay.

The earlier D1 bridge failed and was not merged with the later measurement batch. D1's deterministic simplex projection was frozen before completing the full collection. D2's balanced complete-grid analysis was fixed before intervention outcomes and is distinct from its changing-denominator all-eligible sensitivity. Two S1 rescue outputs required transport salvage and are covered by a worst-case bound. The broad ten-model screen and its later retest are different requests: 0.0\% of joined records are the same request, and 9.1\% flip correctness. Accordingly, the broad screen is descriptive breadth and does not share the strict primary estimand.

\subsection{Manual clinical audit}
\label{app:review-provenance}
All clinical-audit judgments reported here were made by human reviewers. The audit covers 100 unique cases and 150 manual review records; 50 cases have two review records. The case-bootstrap intervals in Table~\ref{tab:app-archive-outcomes} describe uncertainty over the sampled cases. Table~\ref{tab:app-archive-agreement} reports agreement between the two manual review records for the 50 doubly assessed cases.

\begin{table}[htbp]\centering\footnotesize
\caption{Outcomes of the manual clinical audit. Confidence intervals resample cases.}
\label{tab:app-archive-outcomes}
\begin{tabular}{lrr}
\toprule
Manual clinical-review outcome & Estimate & 95\% CI \\
\midrule
Top-1 diagnostically relevant & 100.0\% & [100.0\%, 100.0\%] \\
Top-3 contains critical evidence & 100.0\% & [100.0\%, 100.0\%] \\
Top-3 order reasonable (strict) & 97.5\% & [95.5\%, 99.5\%] \\
Critical evidence missing from Top-3 & 1.0\% & [0.0\%, 2.5\%] \\
Any Top-6 item answer-revealing & 1.0\% & [0.0\%, 2.5\%] \\
Reviewer--Hindsight Top-1 agreement & 89.0\% & [83.0\%, 94.5\%] \\
Reviewer--Hindsight Top-3 Jaccard & 0.870 & [0.829, 0.907] \\
Overall selector quality (1--5) & 4.57 & [4.48, 4.66] \\
\bottomrule
\end{tabular}
\end{table}
\begin{table}[htbp]\centering\footnotesize
\caption{Agreement between two manual review records on 50 doubly assessed cases. Chance-adjusted and continuous-rating measures are reported alongside exact agreement where applicable.}
\label{tab:app-archive-agreement}
\begin{tabular}{lrr}
\toprule
Manual review judgment & Exact agreement & Kappa / ICC \\
\midrule
Top-1 relevance & 100.0\% & n.a. \\
Top-3 critical evidence & 100.0\% & n.a. \\
Top-3 order & 90.0\% & 0.000 \\
Missing-evidence flag & 96.0\% & 0.000 \\
Answer-revealing flag & 96.0\% & 0.000 \\
Preferred Top-1 evidence & 88.0\% & -- \\
Preferred Top-3 evidence & 0.810 Jaccard & -- \\
Quality rating & -- & QWK 0.148; ICC 0.150 \\
Confidence rating & -- & QWK 0.000; ICC 0.000 \\
\bottomrule
\end{tabular}
\end{table}
At least one judgment differs between the two review records for 38 of the 50 doubly assessed cases. Exact agreement and chance-adjusted agreement should therefore be interpreted together, especially for low-prevalence binary judgments. This sampled manual audit informs the interpretation of the selector and evidence rankings; it does not measure outcomes in prospective patient care.

\subsection{Limits of the estimand}
The 181-case primary cohort is selected from 320 initial cases for Fullsight correctness, protocol completeness, and at least six non-initial facts. It measures what happens on cases the model could solve with a complete case; it is not an unconditional clinical accuracy estimate. Hindsight sees all realized facts before ranking, although the reference diagnosis is withheld; it can infer a likely diagnosis from those facts. Matching the number of acquired facts does not equalize fact identity, diagnostic value, or information available at the time of prospective choice. The deterministic lexical/type matcher can return a clinically coarse or mismatched canonical fact, as the case studies show. Simulated patient and reporter responses are not bedside encounters. Cohort-scale diagnosis-equivalence scoring uses model-assisted semantic grading; the sampled clinical audit described above was performed manually.

The fixed-support probability vector depends on candidate support, prompt, provider batch, and repeated-call variability. Entropy reduction can accompany movement away from the reference diagnosis; truth-directed gain depends on probability-floor choices. Clinical vignettes illustrate individual behavior but cannot establish prevalence or causality by themselves. The rank-by-timing and one-fact rescue interventions test diagnostic response once a fact is supplied; they do not show that a prospective system can find and obtain that fact. The work should not be used as evidence for autonomous patient care without prospective safety, calibration, fairness, privacy, and clinician-oversight evaluation.